\documentclass[10pt,twocolumn,letterpaper]{article}

\usepackage{iccv}              

\definecolor{iccvblue}{rgb}{0.21,0.49,0.74}
\usepackage[pagebackref,breaklinks,colorlinks,allcolors=iccvblue]{hyperref}

\usepackage{booktabs}       
\usepackage{graphicx}
\usepackage{float}
\usepackage{array}
\usepackage{soul}
\usepackage{multirow}
\usepackage[accsupp]{axessibility}  
\usepackage{cuted}
\usepackage{babel}

\newtheorem{definition}{Definition}
\newtheorem{proposition}{Proposition}

\newtheorem{remark}{Remark}

\newcommand*\samethanks[1][\value{footnote}]{\footnotemark[#1]}

\def\paperID{8210} 
\def\confName{ICCV}
\def\confYear{2025}

\title{Scendi Score: Prompt-Aware Diversity Evaluation via \\ Schur Complement of CLIP Embeddings}

\author{Azim Ospanov\thanks{The Chinese University of Hong Kong, Department of Computer Science \& Engineering}\\
{\tt\small aospanov9@cse.cuhk.edu.hk}
\and
Mohammad Jalali\samethanks\\
{\tt\small  mjalali24@cse.cuhk.edu.hk}
\\
\and
Farzan Farnia\samethanks\\
{\tt\small  farnia@cse.cuhk.edu.hk}
}

\begin{document}

\maketitle
\vspace{-10mm}
\begin{strip}
\centering
\includegraphics[width=0.92\linewidth]{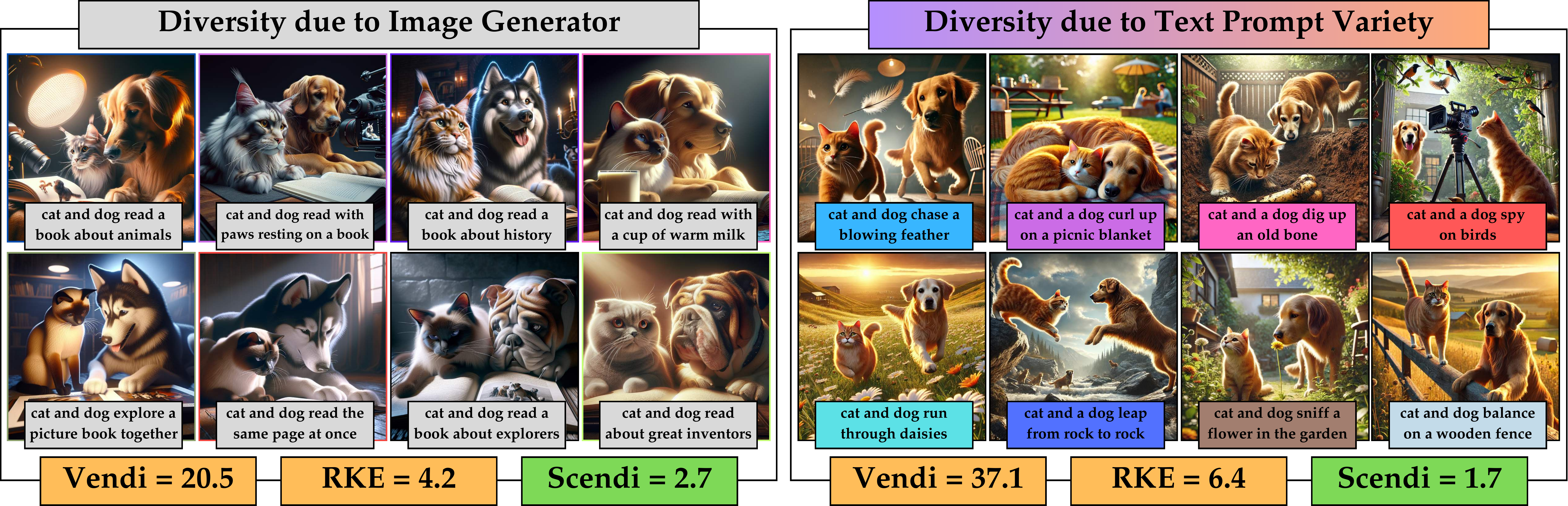}
\captionof{figure}{Comparison of model-driven diversity (left) and prompt-driven diversity (right) in 200 images generated for similar "cat and dog" prompts (left) vs. 200 images generated for diverse "cat and dog" prompts (right). The diversity metrics for unconditional generation, Vendi and RKE, favor the right-side case, but the Scendi score captures intrinsic model diversity, suggesting the left side has higher diversity.}
\label{fig:model vs prompt diversity}
\end{strip}

\begin{abstract}
The use of CLIP embeddings to assess the fidelity of samples produced by text-to-image generative models has been extensively explored in the literature. While the widely adopted CLIPScore, derived from the cosine similarity of text and image embeddings, effectively measures the alignment of a generated image, it does not quantify the diversity of images generated by a text-to-image model. In this work, we extend the application of CLIP embeddings to quantify and interpret the intrinsic diversity of text-to-image models, which are responsible for generating diverse images from similar text prompts, which we refer to as prompt-aware diversity.  To achieve this, we propose a decomposition of the CLIP-based kernel covariance matrix of image data into text-based and non-text-based components. Using the Schur complement of the joint image-text kernel covariance matrix, we perform this decomposition and define the matrix-based entropy of the decomposed component as the \textit{Schur Complement ENtopy DIversity (Scendi)} score, as a measure of the prompt-aware diversity for prompt-guided generative models. Additionally, we discuss the application of the Schur complement-based decomposition to nullify the influence of a given prompt on the CLIP embedding of an image, enabling focus or defocus of the embedded vectors on specific objects. We present several numerical results that apply our proposed Scendi score to evaluate text-to-image and LLM (text-to-text) models. Our numerical results indicate the success of the Scendi score in capturing the intrinsic diversity of prompt-guided generative models. The codebase is available at \url{https://github.com/aziksh-ospanov/scendi-score}.

\end{abstract}

\section{Introduction}

Prompt-guided generative models, which generate data guided by an input text prompt, have gained significant attention in the computer vision community. In particular, text-to-image and text-to-video models, which create visual data based on input text, have found many applications and are widely used across various content creation tasks. Given the important role of prompt-guided generative AI models in numerous machine learning applications, their training and evaluation have been extensively studied in recent years. A comprehensive evaluation of these models, addressing fidelity and diversity, is essential to ensure their effectiveness and adaptability across different use cases.

As the CLIP model \cite{pmlr-v139-radford21a} provides a joint representation for text and image data, the CLIP embeddings have been widely used to evaluate and interpret the performance of text-to-image models. By calculating the cosine similarity between the CLIP embeddings of text and image data, the CLIPScore \cite{hessel2021clipscore} serves as a fidelity metric for measuring the alignment between the text and the generated image. While the CLIPScore and similar uses of CLIP embeddings focus on evaluating the quality of generated samples, these embeddings have not yet been applied to assess the diversity of data produced by text-to-image models. Existing evaluation frameworks typically address diversity scores for unconditional sample generation (considering no input prompt), such as Recall \cite{sajjadi2018assessing,kynkaanniemi2019improved}, Coverage \cite{naeem2020reliable}, RKE \cite{jalali2023information}, and Vendi \cite{friedman2022vendi}, which assess image samples independently of the input text.

However, the diversity of images generated by a text-to-image model depends both on the variety of input text data and on the model’s intrinsic diversity, driving the model to produce varied images in response to similar prompts. Therefore, existing diversity metrics for unconditional sample generation cannot differentiate between diversity arising from varied prompts, i.e., \emph{prompt-driven diversity}, and diversity contributed by the model itself, i.e., \emph{model-driven diversity}. In this work, we leverage CLIP embeddings to propose a framework for quantifying and interpreting the intrinsic diversity of both text-to-image and image captioning (image-to-text) models. The primary objective of our approach is to decompose the CLIP image embedding into a text-based component, influenced by the input text, and a non-text-based component, arising from the model’s inherent randomness.

To achieve this goal, we extend the kernel matrix entropy measures, i.e., Vendi and RKE scores, to enable a prompt aware diversity measurement. To do this, we consider a kernel similarity function and focus on the kernel covariance matrix for the generated (text, image) pairs. Considering the matrix-based entropy of
the Image-based kernel covariance component $C_{II}$, one obtains the existing prompt-unaware Vendi \cite{friedman2022vendi} and RKE \cite{jalali2023information} scores. In this work, we propose
 applying the Gaussian elimination approach, and then decompose the image sub-covariance matrix, \( C_{II} \) as follows:
\begin{equation*}
    C_{II} = \underbrace{C_{II} - C_{IT}C_{TT}^{^{\text{\scriptsize $-1$}}}C^{^{\text{\scriptsize $\top$}}}_{IT}}_{\text{\rm Model-driven Covariance Component}} + \underbrace{C_{IT}\, C_{TT}^{^{\text{\scriptsize $-1$}}}\, C_{IT}^{^{\text{\scriptsize $\top$}}}}_{\text{\rm Prompt-driven Covariance Component}} 
\end{equation*}
Note that the above is the sum of the \emph{Schur~Complement} component, \( C_{II} - C_{IT}C_{TT}^{-1}C_{IT}^\top \), which represents the model-driven diversity component, and the remainder term \( C_{IT}C_{TT}^{-1}C_{TI} \), which captures the prompt-driven variety in the generated images. Extending the Vendi and RKE score to prompt-aware diversity evaluation, we define the matrix-based entropy of the Schur complement component, as \emph{Schur~Complement ENtropy DIversity (Scendi)} score. We highlight that Scendi is a measure of model-driven diversity in the prompt-guided generated data. 

\begin{figure}[t]
    \centering
    \includegraphics[width=\linewidth]{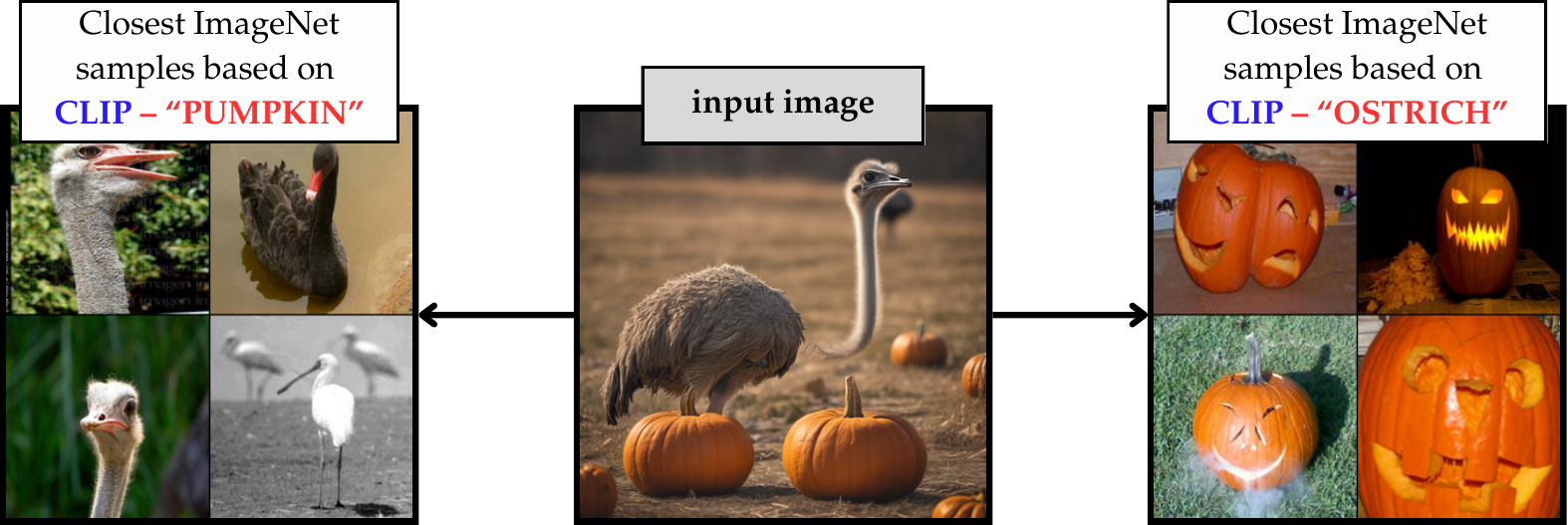}
    \caption{Example of the CLIP embedding decomposition for an input image of "Pumpkin next to Ostrich" generated by SDXL~\cite{podell2024sdxl}. ImageNet~\cite{deng2009imagenet} samples with the highest CLIP similarity to the input image after removing the “pumpkin” direction from its CLIP embedding via the Schur complement (left) and after removing the “ostrich” direction (right).}
    \label{fig:ostrich and pumpkin interpretability}
    \vspace{-5mm}
\end{figure}

To illustrate the effectiveness of the Scendi score in distinguishing model-driven diversity from prompt-driven effects, we present a comparative example. Figure~\ref{fig:model vs prompt diversity} compares the Scendi diversity scores for two sets of samples generated by DALL-E 3 \cite{dalle3}. The left-side subfigure is designed to represent model-driven diversity, showing images of diverse-breed cats and dogs generated using similar prompts like “cat and dog read a text.” In contrast, the right-side subfigure highlights the effects of prompt-driven diversity, featuring same-breed cats and dogs across diverse-occasion prompts. The unconditional diversity metrics, Vendi and RKE, report higher diversity for the right-side case. In contrast, our proposed Scendi score, designed to isolate intrinsic model diversity from prompt-driven effects, indicates greater diversity in the left-side case. This example highlights the advantage of the prompt-aware diversity measurement by Scendi score over existing unconditional diversity metrics in capturing the model-driven diversity.  


We present several numerical applications of the proposed framework to evaluate and interpret standard text-to-image and image-captioning models. Our results on several simulated scenarios with known ground-truth diversity indicate that the proposed entropy metric correlates with the non-text-based diversity in images, capturing variation not attributable to the text prompt. Additionally, we show that the decomposed feature component can neutralize the influence of specific objects in the text prompt within the image embedding. Specifically, we use this decomposed feature to diminish the impact of visible text in images, reducing its effect on the embedding. We also demonstrate how the decomposed image embedding can enhance or reduce the focus on particular objects or styles within an image, which can have implications for downstream applications of CLIP embeddings when emphasizing specific elements. For instance, Figure~\ref{fig:ostrich and pumpkin interpretability} shows a generated image of an ostrich next to a pumpkin. We showcase the ImageNet samples that achieve the highest CLIPScore when compared to the modified image embeddings obtained by canceling the “pumpkin” and “ostrich” directions. This demonstrates that the modified embeddings effectively eliminate the influence of those unwanted objects. The main contributions of this work are summarized as follows:
\begin{itemize}[leftmargin=*]
    \item Proposing a Schur Complement-based approach to text-to-image diversity evaluation that decomposes the diversity metric into prompt-induced and model-induced components
    \item Providing a decomposition method that allows to remove directions from the CLIP embedding based on the Schur Complement modified image embedding
    \item Presenting numerical results on the Schur complement-based decomposition of CLIP embeddings and Scendi score performance under various diversity scenarios and data modalities.
\end{itemize}

\section{Related Work}

\noindent\textbf{CLIP interpretability and decomposition.} Contrastive vision-and-language models, such as CLIP \cite{pmlr-v139-radford21a} are a class of models that were trained on paired text prompts and images. The notable feature of CLIP is a shared embedding space between image and text data. 
A common interpretability method involves heatmaps to highlight relevant image areas \citep{gradcam,integratedgradients,Chefer_2021_CVPR,gong2025boosting}. However, heatmaps are used to identify objects and lack spatially dependent information, such as object size and embedding output. Other approaches require decomposing the model architecture and analyzing attention heads. \cite{gandelsman2024interpreting} introduced \emph{TextSpan}, which finds a vector direction for each attention head and assigns it an appropriate text label. Another prominent approach, proposed in \cite{bhalla2024interpretingclipsparselinear}, decomposes dense CLIP embeddings into sparse, interpretable semantic concepts to enhance embedding description. \cite{materzynska2022disentangling} suggests disentangling written words from underlying concepts via orthogonal projection of learned CLIP representations. \cite{brack2023sega} utilize semantic guidance to move between concepts and introduce edits during diffusion process. Recently, \cite{levi2025the} studied the geometry of the CLIP embeddings and identified numerous properties such as modality mass of image and text components in the embedding space.

\noindent\textbf{Evaluation of Diversity in Generative Models.} Diversity evaluation in generative models has been extensively explored in the literature, with existing metrics broadly classified as either reference-based or reference-free. Reference-based metrics, such as FID \cite{heusel2017gans}, KID \cite{binkowski2018demystifying}, and IS \cite{Salimans2016}, assess diversity by quantifying the distance between true and generated data. Metrics like Density and Coverage \cite{naeem2020reliable}, and Precision and Recall \cite{kynkaanniemi2019improved} evaluate quality and diversity by analyzing the data manifold. Reference-free metrics, including Vendi \cite{friedman2022vendi, pasarkar2023cousins}, RKE \cite{jalali2023information}, and their improved and convergent variants FKEA-Vendi \cite{ospanov2024towards} and Truncated Vendi~\cite{ospanov2025do}, measure the entropy of a similarity metric (e.g., kernel matrix),  capturing the number of distinct modes in the data. Since these metrics do not require a reference distribution, they are particularly suitable for text-to-image model evaluation, where selecting an appropriate reference dataset is challenging. Another concurrent work on prompt-aware diversity is the Conditional Vendi score that measures the entropy of the Hadamard kernel product subtracted by the text component.~\cite{jalali2024conditionalvendiscoreinformationtheoretic}. Also, the kernel-based evaluation scores have been further utilized in the context of online model selection \cite{hu2025multiarmedbanditapproachonline,rezaei2025diversediverseoptimalmixtures,hu2025online,hu2025promptwiseonlinelearningcostaware}, novelty evaluation and detection \cite{zhang2024interpretable,zhang2025unveiling}, distributed evaluation \cite{wang2023distributed}, and embedding comparison and alignment \cite{jalali2025towards,gong2025kernel}.

\noindent\textbf{Evaluation of Text-to-Image Generative Models.} Conditional generation models have been extensively studied, with CLIPScore \cite{hessel2021clipscore} and its variations, such as Heterogeneous CLIPScore \cite{kim2024attribute}, being the standard metrics for assessing prompt-image alignment using cosine similarity. Another key approach is the FID framework adapted for conditional models, which measures the joint distribution distance between prompts and images, known as FJD \cite{devries2019evaluationconditionalgans}. Holistic evaluation methods, such as the HEIM \cite{lee2023holistic} and HELM \cite{liang2023holistic} benchmarks, unify different aspects of generated data to provide comprehensive assessments. Diversity evaluation typically involves generating multiple images per prompt and measuring their inter-diversity \cite{astolfi2024consistencydiversity, kannen2024aesthetics}. Existing metrics require redundant image generation for diversity measurement, whereas our proposed Schur Complement-based decomposition bypasses this need, enabling evaluation on pre-generated datasets without requiring multiple images per prompt.

\section{Preliminaries}


\noindent\textbf{Kernel Functions.} We call $k:\mathcal{X}\times \mathcal{X} \rightarrow \mathbb{R}$ a kernel function if for samples $x_1, \ldots, x_n \in \mathcal{X}$, the resultant kernel matrix $K = \bigl[k(x_i,x_j) \bigr]_{1\le i,j\le n}$ is a PSD (positive semi-definite) matrix. Moreover, $K$ can be decomposed into a dot product of the kernel feature maps $\phi:\mathcal{X} \rightarrow \mathbb{R}^d$ as follows:
\begin{equation}\label{Eq: Kernel Equivalent Definition}
    k(x,x') \, =\,  \bigl\langle \phi(x), \phi(x') \bigr\rangle 
\end{equation}
In this work, we provide several numerical results for the cosine-similarity Kernel function defined as:
\begin{equation}
    k_{\text{\rm Cosine-Similarity}} \bigl(x , x'\bigr) \, :=\, \frac{\bigl\langle x, x' \bigr\rangle}{\Vert x\Vert_2\Vert x'\Vert_2}
\end{equation}
Also, we consider the Gaussian Kernel defined as:
\begin{equation}
    k_{\text{\rm Gaussian}(\sigma)} \bigl(x , x'\bigr) \, :=\, \exp\Bigl(-\frac{\bigl\Vert x - x'\bigr\Vert^2_2}{2\sigma^2}\Bigr)
\end{equation}
Note that both of the presented kernels are normalized kernels, since  $k(x,x)=1$ holds for every $x\in\mathcal{X}$. To apply entropy measures, we consider the normalized kernel matrix given by $\frac{1}{n}K$. Therefore, we observe $\mathrm{Tr}(\frac{1}{n}K)=1$ for every normalized kernel, implying that the eigenvalues of $\frac{1}{n}K$ form a probability model, since they are non-negative and sum up to $1$. Note that $\mathrm{Tr}(\cdot)$ denotes the matrix trace.

\noindent\textbf{Matrix-based Entropy and Kernel Covariance Matrix.} Given a PSD matrix $A$ with unit trace and eigenvalues $\lambda_1,\hdots,\lambda_n \ge 0$, the Von-Neumann entropy is defined as:\vspace{-2mm} 
\begin{equation}
    H(A) := \sum_{i=1}^n\lambda_i \log \frac{1}{\lambda_i}.
\end{equation}
\cite{friedman2022vendi} discusses that the Von-Neumann entropy of the normalized kernel matrix $\frac{1}{n}K$ can effectively capture the entropy of the cluster variable in the collected data, and proposes the Von Neumann entropy diversity (Vendi) score for measuring the diversity of a sampleset. Observe that $\frac{1}{n}K$ shares the same non-zero eigenvalues with the kernel covariance matrix $C_X$ defined as:
\begin{equation}
    C_X := \frac{1}{n}\sum_{i=1}^n \phi(x_i)\phi(x_i)^\top  = \frac{1}{n} \Phi^\top \Phi
\end{equation}
where $\Phi\in\mathbb{R}^{n\times d}$ is an $n\times d$ matrix whose rows are the feature presentations of samples $\phi(x_1),\ldots,\phi(x_n)$. Therefore, given the eigenvalues $\lambda_1,\ldots ,\lambda_d$ of the kernel covariance matrix $C_X$, the Vendi and RKE \cite{jalali2023information} scores are\vspace{-4mm}
\begin{align*}
    \mathrm{Vendi}(x_1,\ldots ,x_n) &= \exp\Bigl( \sum_{i=1}^d \lambda_i \log\frac{1}{\lambda_i}\Bigr), \\
    \mathrm{RKE}(x_1,\ldots ,x_n) &= \frac{1}{\sum_{i=1}^d \lambda_i^2} = \frac{1}{\Vert C_X\Vert^2_F}.
\end{align*}
Feature representation varies between kernel methods. In Cosine Similarity Kernel, $\phi(x) = \text{CLIP}(x) / \Vert 
\text{CLIP}(x) \Vert_2$, i.e. normalized CLIP embedding of sample $x$, whereas in shift-invariant kernels, $\phi$ is a proxy feature map of the Gaussian kernel following the random Fourier features \cite{rahimi2007random}.

\noindent\textbf{Schur Complement.} Consider a block matrix $\Lambda$ with a symmetric matrix partition as follows:
\[
\Lambda = \begin{bmatrix} B & C \\ C^\top & D \end{bmatrix}
\]
where \( B \) and \( D \) are square symmetric submatrices. If \( B \) is invertible, the Schur Complement of \( B \) in \( A \) is given by $
S = D - C^\top B^{-1} C $. 
In general, even if $B$ is not invertible, $B^{-1}$ can be replaced with the Moore-Penrose psuedoinverse $B^\dagger$ in the Schur complement definition. Note that the Schur complement $S\succeq \mathbf{0}$ will be PSD for a PSD matrix $\Lambda \succeq \mathbf{0}$. 

\begin{figure*}[h]
    \centering
    \includegraphics[width=\linewidth]{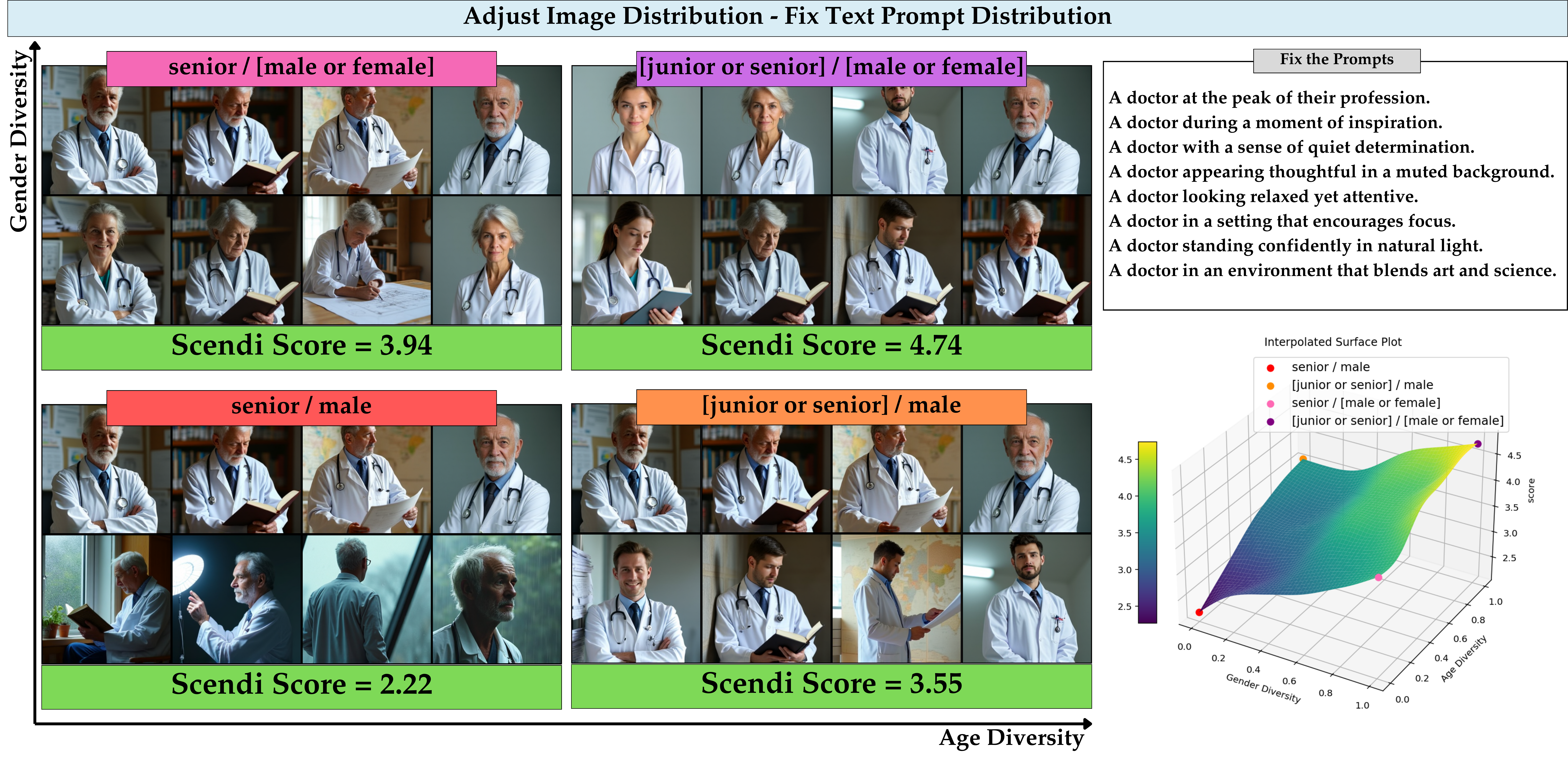}
    \caption{Evaluated $\mathrm{Scendi}$ scores with fixed text prompt distribution about doctors in various settings. Interpolated surface plot in the bottom right visualises the change in diversity (z-axis) when doctors have a varying diversity of age and gender features. (x and y axis)}
    \label{fig:gender age doctors fixed prompts}
\end{figure*}

\section{Diversity Evaluation for Text-to-Image Generative Models via CLIP Embedding}
As discussed earlier, the CLIP model offers a joint embedding of text and image data, which enables defining joint (text,image) kernel covaraince matrices for the collected data. Suppose that we have collected $n$ paired text,image samples $(T_j,I_j)$ for $j=1,\ldots ,n$. Here $I_j$ represents the $j$th image and $T_j$ represents the corresponding text. The application of CLIP embedding transfers the pair to the share embedding space $\mathcal{X}$ and results in embedded samples $(x_{T_j}, x_{I_j}) $ in the CLIP space. As a consequence of the joint embedding, not only can we compute the kernel function between (text,text) and (image,image) pairs, but also we can compute the kernel function for a (text,image) input.

To analyze the embedded sample, consider a kernel function $k: \mathcal{X} \times \mathcal{X}\rightarrow \mathbb{R}$ with feature map $\phi:\mathcal{X}\rightarrow \mathbb{R}^d$ where $\mathcal{X}$ is the CLIP space and $d$ is the dimension of the kernel feature map. Then, for the collected samples, we define the embedded image feature matrix $\Phi_I\in\mathbb{R}^{n\times d}$ whose $j$th row is $\phi(x_{I_j})$ for the $j$th image. Similarly, we define the embedded text feature matrix $\Phi_T\in\mathbb{R}^{n\times d}$ for the text samples. Note that the resulting CLIP-based kernel covariance matrix for the joint (text,image) map $[\phi(\mathbf{x}_T),\phi(\mathbf{x}_I)]$ is
\begin{equation*}
    C_{\text{\rm joint (I,T)}} := \begin{bmatrix} C_{II}& C_{IT}\vspace{2mm} \\  C_{IT}^{^{\text{\scriptsize$\top$}}} & C_{TT}\vspace{1mm} \end{bmatrix}
\end{equation*}
In the above, we define the sub-covariances as follows:
\begin{equation*}
    C_{II} = \frac{1}{n} \Phi_I^\top \Phi_I \, , \, C_{IT} = \frac{1}{n} \Phi_I^\top \Phi_T \, , \, C_{TT} = \frac{1}{n} \Phi_T^\top \Phi_T
\end{equation*}
Note that the above matrix is PSD, which implies that we can leverage the Schur complement to decompose the image-based block $C_{II}$ as follows:
\begin{equation}\label{Eq: Schur component decomposition}
    C_{II} = \underset{\text{\normalsize model-driven component $\Lambda_I$}}{\underbrace{C_{II} - C_{IT}C_{TT}^{-1}C^\top_{IT}}} + \underset{\text{\normalsize text component $\Lambda_T$}}{\underbrace{C_{IT}C_{TT}^{-1}C^\top_{IT}}}
\end{equation}\vspace{-3mm}
\begin{proposition}\label{Prop: 1}
Define text-to-image conversion matrix $\Gamma^* = C_{IT}C_{TT}^{-1}  $. Then, $\Gamma^*$ is an optimal solution to the following ($\Vert\cdot\Vert_F$ denotes the Frobenius norm):
\begin{equation*}
    \underset{\Gamma\in\mathbb{R}^{d\times d}}{\arg\!\min}\;\; \frac{1}{n}\Bigl\Vert \Phi_I^\top - \Gamma\Phi_T^\top \Bigr\Vert_F^2
\end{equation*}
Then, in \eqref{Eq: Schur component decomposition}, $\Lambda_T$ is the covariance matrix of $\Gamma^*\phi(x_T)$, and $\Lambda_I$ is the kernel covariance matrix of $\phi(x_I)-\Gamma^*\phi(x_T)$.
\end{proposition}
\begin{remark}
Proposition \ref{Prop: 1} shows that given the optimal text-to-image conversion matrix $\Gamma^*= C_{IT}C_{TT}^{-1}$, the effect of a text $T$ on the embedding of an image $I$ can be canceled by considering the remainder term $\phi(x_I)- \Gamma^*\phi(x_T)$ to decorrelate the image and text embedding. 
\end{remark}
If we consider the cosine-similarity kernel feature map $\phi(x)=x/\Vert x\Vert$, the above discussion reveals an approach to cancel the effect of the input text on the CLIP embedding of the output images. Given a paired dataset of prompts and images, we first compute the modification matrix $\Gamma^* = C_{IT}C_{TT}^{-1}$ and modify the CLIP embedding for the input image  $x_I=\mathrm{CLIP}(I)$ and prompt $x_T= \mathrm{CLIP}(T)$ as:
\begin{equation}
    \mathrm{CLIP}_{\mathrm{modified}}(I | T) := \mathrm{CLIP}(I) - \Gamma^*\mathrm{CLIP}(T) 
\end{equation}
Note that using the cosine similarity kernel, the dimension of $\Gamma^*$ matches with the CLIP dimension, i.e., 512. 

\begin{figure*}[ht]
    \centering
    \includegraphics[width=\linewidth]{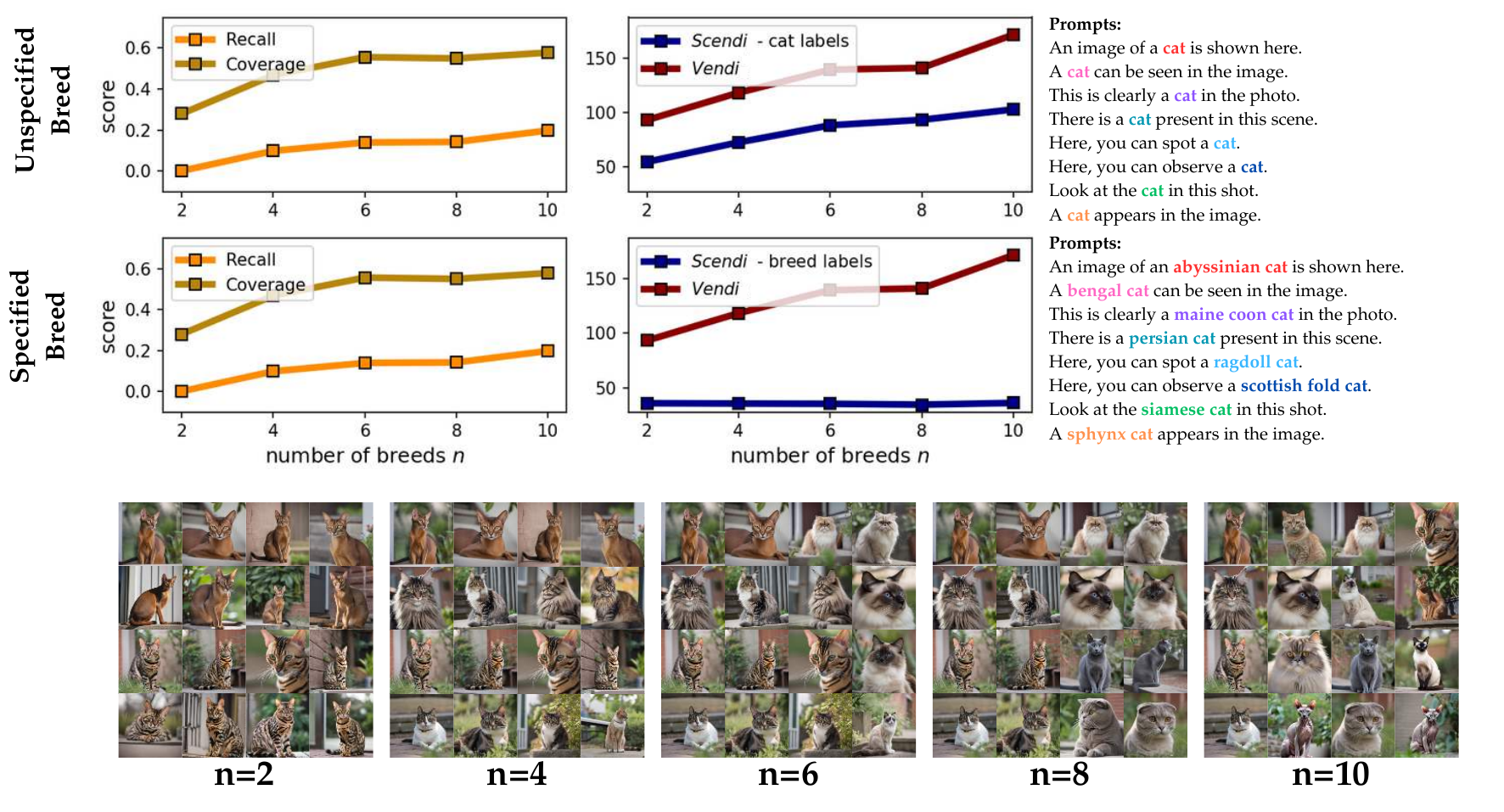}
    \caption{Evaluated $\mathrm{Scendi}$, Vendi, Recall and Coverage scores with Gaussian Kernel on different cat breeds dataset.}
    \label{fig:cat breed gaussian diversity plots}
    \vspace{-2mm}
\end{figure*}
\begin{figure}[ht]
    \centering
    \includegraphics[width=\linewidth]{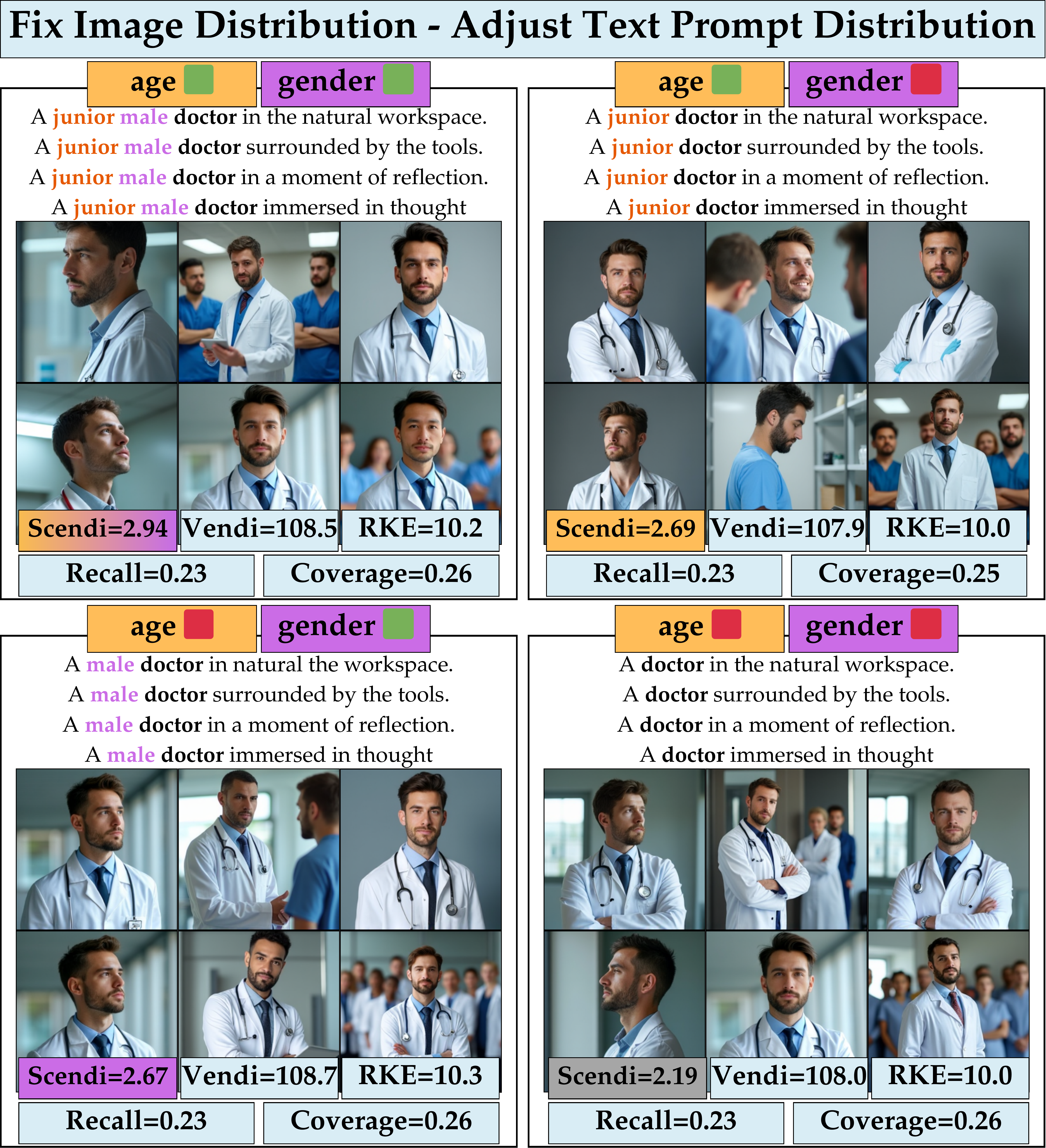}
    \caption{Evaluated $\mathrm{Scendi}$ scores with fixed text prompt distribution. $\mathrm{Scendi}$ changes to reflect the diversity contributed by the varying level of information in text prompts.}
    \label{fig:gender age doctors fixed images}
    \vspace{-5mm}
\end{figure}

Furthermore, in the decomposition in \eqref{Eq: Schur component decomposition}, both the model-driven covariance component $\Lambda_I$ and the text prompt-driven covariance component $\Lambda_T$ are PSD matrices, which have non-unit trace values. To apply the matrix-based entropy definition which requires the unit trace, we rewrite the identity as 
\begin{equation*}
   C_{II} = \mathrm{Tr}(\Lambda_I)\cdot \frac{1}{\mathrm{Tr}(\Lambda_I)}\Lambda_I + \bigl(1-\mathrm{Tr}(\Lambda_I)\bigr)\cdot \frac{1}{\mathrm{Tr}(\Lambda_T)}\Lambda_T 
\end{equation*}

\begin{definition}
We define the Schur-Complement-ENtropy Diversity ($\mathrm{Scendi}$) score 
as follows:
\begin{align*}
    \mathrm{Scendi}(x_1,.. ,x_n;t_1,..,t_n)\, :=& \, \exp \Bigl( \sum_{j=1}^d\lambda^{(\Lambda_I)}_j \log \frac{\mathrm{Tr}(\Lambda_I)}{\lambda^{(\Lambda_I)}_j} \Bigr )
\end{align*}
where $\lambda^{(\Lambda_I)}_j$ denotes the $j$th eigenvalue of matrix $\Lambda_I$ and $  \mathrm{Tr}(\Lambda_I)= \sum_{j=1}^d \lambda^{(\Lambda_I)}_j$ is the sum of the eigenvalues. 
\end{definition}

It can be seen that the defined Schur complement entropy is the conditional entropy of the hidden image cluster variable $\mathrm{Mode}\in\{1,2,\ldots,d\}$, which follows the spectral decomposition of the image covariance matrix $C_{II}=\sum_{j=1}^d \lambda_j v_jv_j^\top$, where the mode $\mathrm{Mode}=i$ has probability $\lambda_i$. Here we define a text-based guess of the image mode $Y_T$, which as we discuss in the Appendix can correctly predict the cluster with probability $\mathrm{Tr}(\Lambda_T)$ and outputs erasure $e$ with probability $\mathrm{Tr}(\Lambda_I)$. Then, we show in the Appendix that following standard Shannon entropy definition:
$$\mathrm{Scendi}(x_1,.. ,x_n;t_1,..,t_n) = \exp\Bigl(H\bigl(\mathrm{Mode} | Y_T\bigr)\Bigr)$$
Therefore, this analysis allows us to decompose the diversity of generated images into model-driven and text-driven components. 

\section{Numerical Results}
We evaluated our proposed Schur Complement-based approach for CLIP decomposition across various scenarios using both real and synthetic datasets. Our experimental results are reported for the standard Cosine Similarity and Gaussian kernel functions. For the Gaussian kernel function, we use random Fourier features \cite{rahimi2007random} with a random Fourier dimension $r=2000$ to embed the kernel using a finite feature dimension. In the diversity evaluation, we report our defined Scendi scores as well as the Vendi score \cite{friedman2022vendi}, RKE\cite{jalali2023information}, Recall\cite{kynkaanniemi2019improved} and Coverage\cite{naeem2020reliable} for unconditional diversity assessment without taking the text data into account. Note that Recall and Coverage may only be performed in the presence of a valid representative reference dataset. The clustering experiments were performed using Kernel PCA with the Gaussian kernel, highlighting the top clusters that align with the top eigenvector directions in the data. 

\begin{figure*}[htp]
    \centering 
    \captionsetup{type=figure}
    \includegraphics[width=\linewidth]{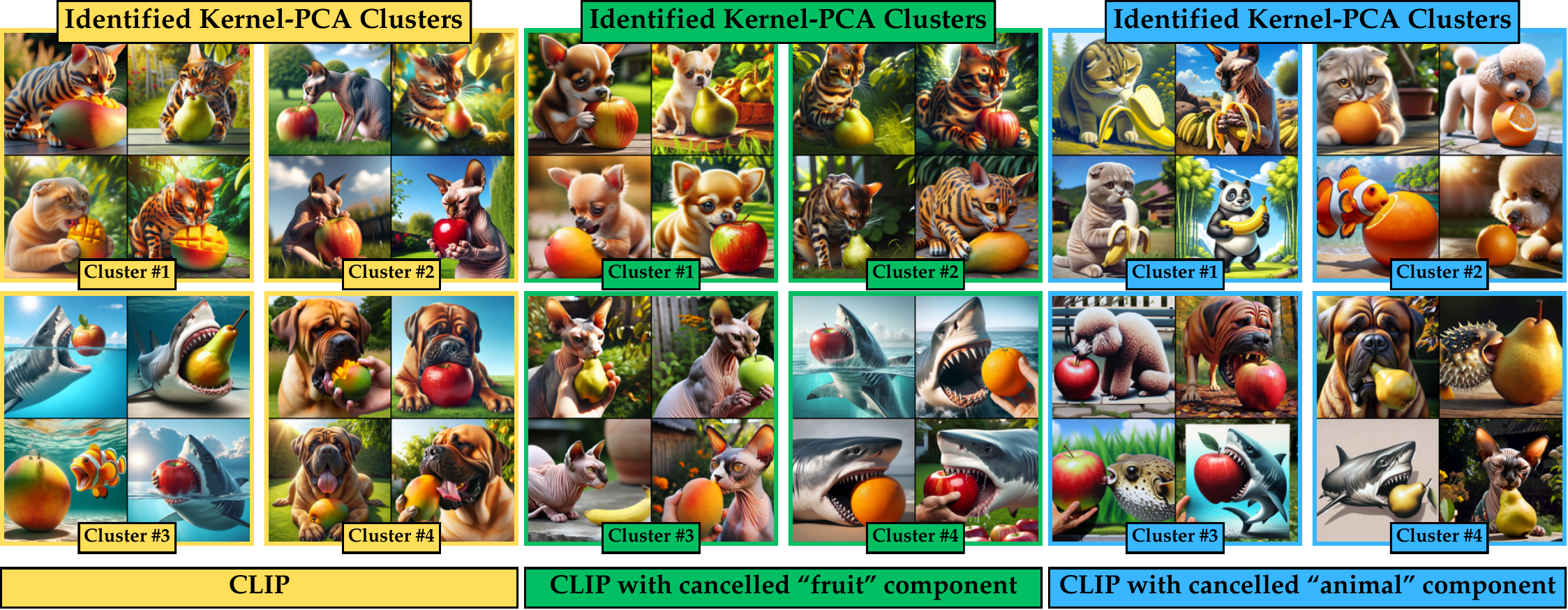}
    \captionof{figure}{Clusters of DALL-E 3 \cite{dalle3} generated images of animals with fruits. The yellow column shows Kernel-PCA (KPCA) clusters using CLIP; the green column shows KPCA clusters with the application the proposed Schur-Complement-based method to remove "fruit" direction from CLIP embedding; and the blue column shows clusters with the "animal" direction removed from CLIP embedding.}
    \label{fig:animals fruits dalle3}
\end{figure*}

\begin{figure}[ht]
  \centering
  \begin{minipage}[t]{0.95\linewidth}
    \centering
    \includegraphics[width=0.95\linewidth]{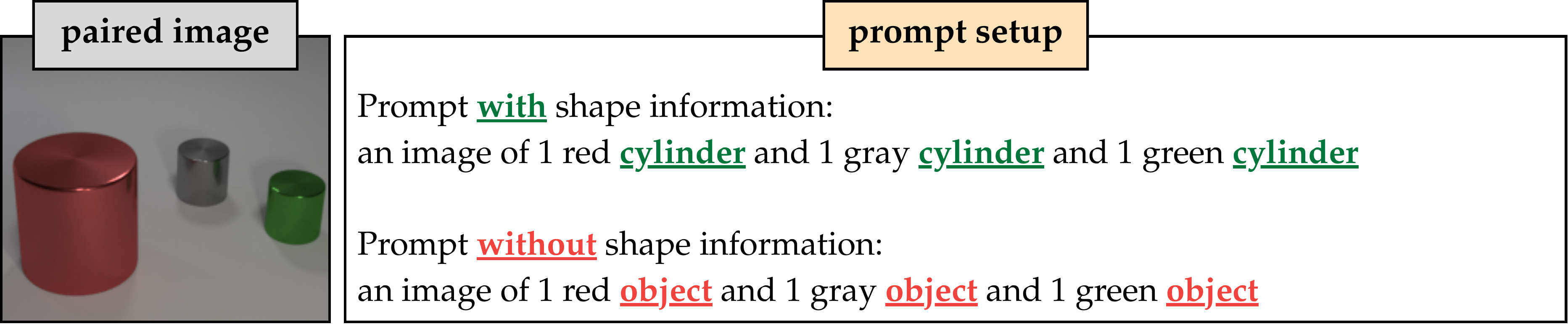} 
    \label{fig:myfig}
  \end{minipage}
  \hfill
  \begin{minipage}[t]{0.48\textwidth}

    \centering
    \vspace{-1mm}
    \resizebox{0.95\linewidth}{!}{%
    \begin{tabular}{lc|ccc}
    \toprule
    metric & prompt setting & 1 shape & 2 shapes & 3 shapes \\
    \midrule
    Vendi                   & –                               & 5.93 & 7.03 & 9.46 \\
    \midrule
    \multirow{2}{*}{$\mathrm{Scendi}$}& w shapes                        & 2.94 & 2.95 & 2.96 \\
                             & w/o shapes                     & 2.53 & 2.96 & 3.70 \\
    \bottomrule
    \end{tabular}
    }
    \vspace{-5mm}
    \label{tab:clevr}

  \end{minipage}

  \vspace{1ex}
  \caption{Diversity measurement on the CLEVR dataset using Vendi and Scendi metrics. Vendi stays constant for all prompts, while $\mathrm{Scendi}$ only increases when shapes aren’t specified, highlighting true image-driven diversity.}
  \label{fig:combined}
\end{figure}

\begin{figure}
    \centering
    \includegraphics[width=\linewidth]{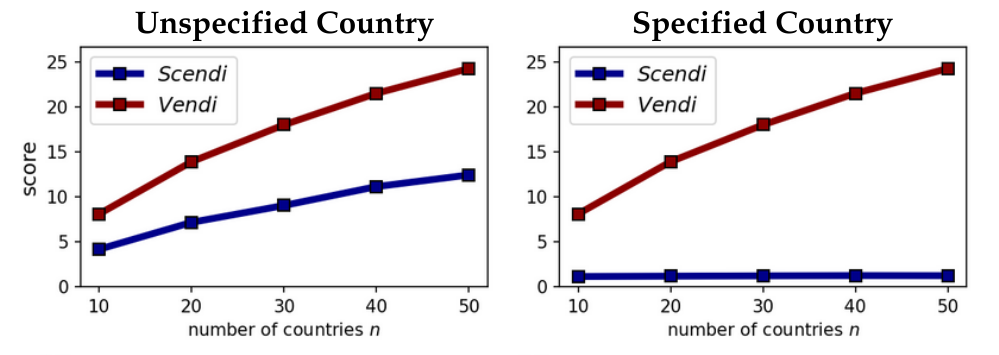}
    \caption{LLM countries with Qwen3-0.6B~\cite{qwen3embedding}}
    \label{fig:llm-countries}
    \vspace{-5mm}
\end{figure}

\textbf{Measuring Text-to-Image Model Diversity.} To highlight the strengths of Scendi-based diversity evaluation, we conducted numerical experiments comparing unconditional and conditional diversity assessments. Figure~\ref{fig:gender age doctors fixed prompts} presents an experiment with the FLUX.1-dev~\cite{flux_2024} model, where individuals are generated based on an ambiguous prompt. We fixed the text prompt distribution and conditioned the model to generate doctors of varying gender and age. The results show that when the model is restricted to producing only "senior male doctors," the $\mathrm{Scendi}$ score is the lowest. Conversely, when the model generates a broader range of age groups and genders, the diversity score increases. We also provide an interpolated surface plot illustrating the intermediate mixing results. A diversity level of 0 indicates a strong bias toward a single characteristic (e.g., all images depict males), whereas a score of 1 signifies an equal representation of all characteristics (e.g., an equal number of male and female images).

\begin{figure*}[h]
    \centering
    \begin{subfigure}[b]{0.48\linewidth}
        \includegraphics[width=\linewidth]{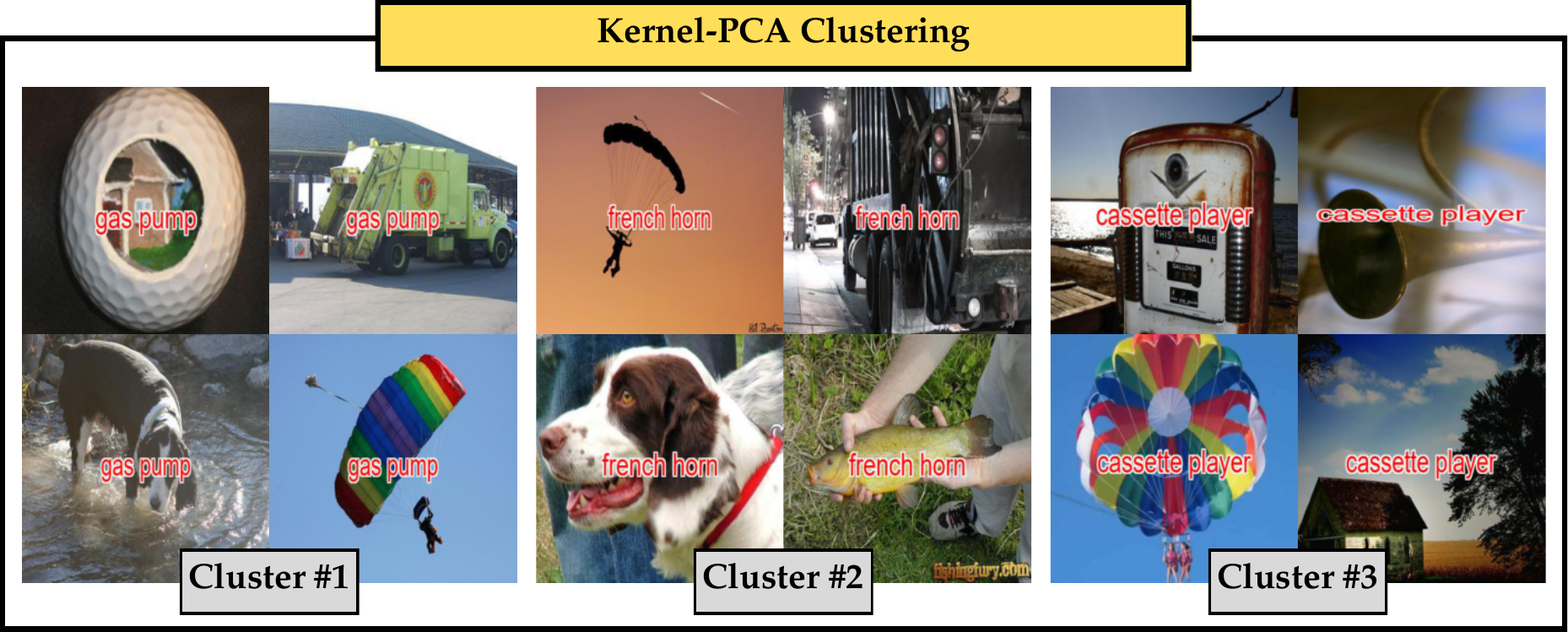}
    \end{subfigure}
    \hfill
    \begin{subfigure}[b]{0.48\linewidth}
        \includegraphics[width=\linewidth]{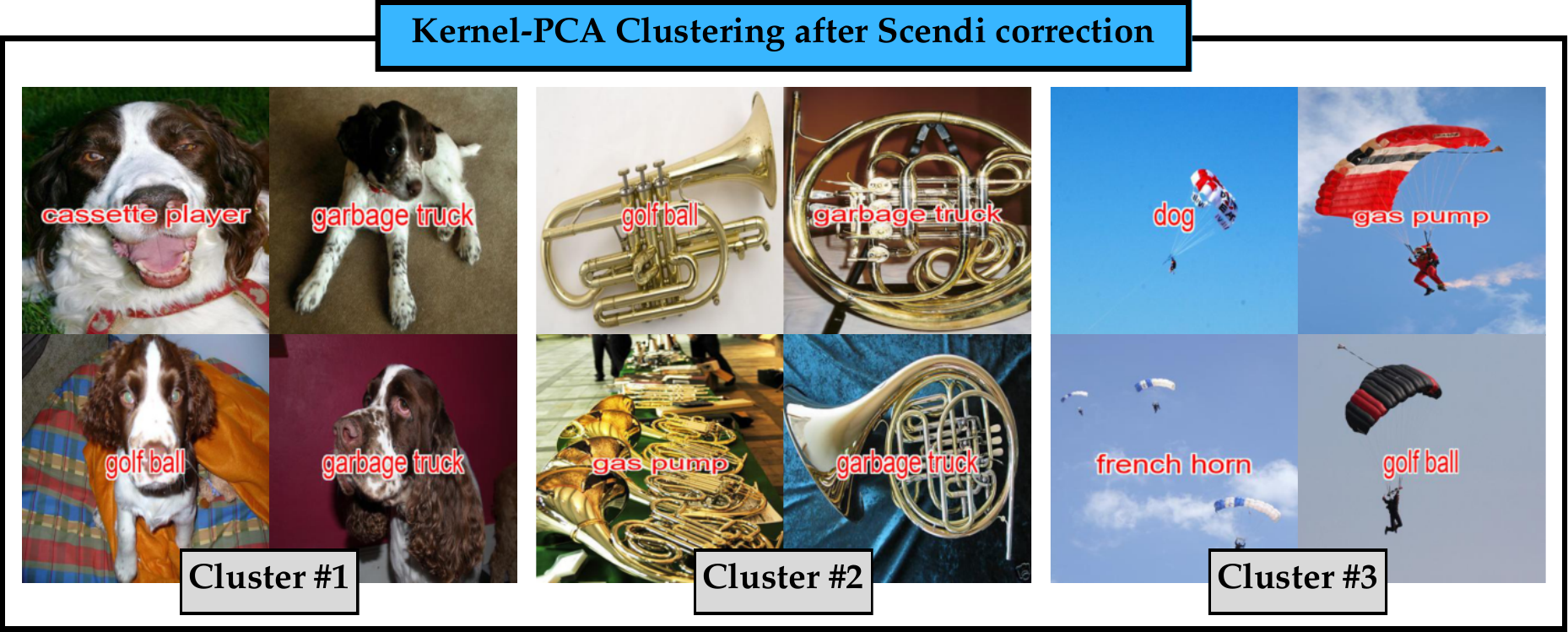}
    \end{subfigure}
    \hfill
    \caption{Kernel PCA clusters before and after CLIP correction on captioned ImageNet dataset.}
    \label{fig:caption clusters}
    \vspace{-2mm}
\end{figure*}

Figure~\ref{fig:gender age doctors fixed images} illustrates the inverse scenario: the image distribution is fixed while the accompanying text prompts vary in their specificity. We generated four sets of "junior male doctor" images with different seeds, while prompts were assigned four levels of detail—fully specified (age and gender), partially specified (age only or gender only), or unspecified (no details). We measured $\mathrm{Scendi}$, Vendi, RKE, Coverage and Recall scores across four scenarios. Reference set for R/C is "doctor" class in IdenProf dataset~\cite{IdenProf}. Vendi, RKE, Recall and Coverage scores remain relatively unchanged, whereas $\mathrm{Scendi}$ distinguishes between these cases. When the description is incomplete, $\mathrm{Scendi}$ is lowest due to the strong presence of "junior male doctors" across all images, indicating bias. However, when prompts explicitly state that all images should depict "junior male doctors," diversity increases and is evaluated based on auxiliary features such as background details rather than age or gender. When partial information is given, $\mathrm{Scendi}$ falls between the fully specified and unspecified cases. These results highlight the need for differentiation between model-driven and prompt-driven diversity. 

To further quantify our findings, in Figure~\ref{fig:cat breed gaussian diversity plots} we conducted an experiment on SDXL generated cat images of different breeds. We generated 1,000 images per class and compared $\mathrm{Scendi}$ with other diversity metrics. When prompts did not specify cat breeds, the diversity captured by $\mathrm{Scendi}$ aligns with unconditional Vendi score. Conversely, when breed information was included in the prompts, diversity mainly stemmed from the text prompts rather than the image generator, leading to a non-increasing $\mathrm{Scendi}$ score. In both scenarios, Vendi, Coverage, and Recall scores remained unchanged. The reference set for Recall and Coverage consisted of ImageNet~\cite{deng2009imagenet} 'cat' samples. Additional experiments on other cases are discussed in the Appendix. We evaluated our metric on a subset of CLEVR~\cite{johnson2016clevrdiagnosticdatasetcompositional} images containing only one shape to benchmark Scendi in a fully controlled environment. Each image was paired with two prompts: one specifying the shape and one omitting it, while keeping object count and colors constant. In a diversity evaluation analogous to Figure \ref{fig:cat breed gaussian diversity plots}, similar trend remains. 

This work primarily focuses on text‑to‑image evaluation using CLIP; however, the Scendi score naturally extends to any uni‑modal evaluation setting, such as text‑to‑text, which has become extremely popular with LLMs. Figure \ref{fig:llm-countries} mirrors our two previous experimental setups, but here the dataset consists of country prompts paired with short, fact‑only responses generated by Deepseek‑V3 \cite{deepseekai2024deepseekv3technicalreport}. Notably, this experiment demonstrates that in a uni‑modal context the Scendi score applies directly and does not require CLIP or other cross-modal embeddings, any suitable embedding will suffice.

\textbf{CLIP Decomposition.} We use the eigenspace of the Schur complement component to interpret the diversity in the generated images that is independent of the text prompt, revealing the unique elements introduced by the generation model in the images. The eigenspace-based interpretation follows the application of the Kernel PCA (KPCA) clustering method and leads to a visualization of the sample clusters due to the input text clusters and the clusters with a feature added by the model to the generated image. Figure~\ref{fig:animals fruits dalle3} shows the resulting KPCA-detected clusters for DALL-E 3 \cite{dalle3} generated images in response to prompts of type "[animal] eating a [fruit]" with 5 different animal and fruit names. The detected clusters of CLIP image embedding (leftmost case) shows clusters with mixed animals and fruits. On the other hand, by considering the prompt of only specifying the fruit, the Schur complement component's KPCA clusters highlight the animals in each cluster (middle case), and similarly by considering the prompt of only specifying the fruit, the animals are captured by each KPCA cluster of the Schur complement matrix. The results show how decomposing and removing the influence of concepts present in the prompts leads to changes in the major clusters. Further decomposition results and analysis are provided in the Appendix.

\textbf{Typographic Attacks.} CLIP’s sensitivity to text within images, as showed in \cite{materzynska2022disentangling}, makes it vulnerable to typographic attacks, where overlaying misleading text on an image influences its classification \cite{lemesle2022languagebiased}. To investigate this, we constructed a dataset of 10 ImageNet classes, each image engraved with a random class label. Figure \ref{fig:caption clusters} demonstrates CLIP's susceptibility, showing that top eigenvector directions capture the engraved text rather than the actual image content. By conditioning on prompts like "text reading 'cassette player' on top of image" and removing this direction, we re-clustered images based on the corrected CLIP embedding. This adjustment allowed us to identify clusters based on the actual image content rather than the engraved text. Additional results are provided in the Appendix.

\section{Conclusion}
In this work, we proposed a Schur Complement-based approach to decompose the kernel covariance matrix of CLIP image embeddings into the sum of text-induced and model-induced kernel covariance components. We demonstrated the application of this Schur Complemental approach to define the Scendi score that evaluates the diversity of prompt-guided generated data in a prompt-aware fashion. This method extends the application of CLIP embeddings from relevance evaluation to assessing diversity in text-to-image generation models. Additionally, we applied our approach to modify CLIP image embeddings by modifying or canceling the influence of an input text. Our numerical results showcase the potential for canceling text effects and its applications to computer vision tasks. Future research directions include extending the Schur Complement-based approach to other generative AI models, such as text-to-video and language models. Additionally, leveraging this decomposition to fine-tune the CLIP model for enhanced understanding of diverse concepts represents another future direction. Finally, applying Scendi score for diversity guidance in diffusion models, similar to RKE guidance in \cite{jalali2025sparke} and Vendi guidance in \cite{hemmat2024improvinggeodiversitygeneratedimages}, will be relevant for future studies. 

\section*{Acknowledgements}
This work is partially supported by a grant from the Research Grants Council of the Hong Kong Special Administrative Region, China, Project 14209920, and is partially supported by CUHK Direct Research Grants with CUHK Project No. 4055164 and 4937054.
Also, the authors would like to thank the anonymous reviewers and meta-reviewer for their constructive feedback and useful suggestions.


{
    \small
    \bibliographystyle{ieeenat_fullname}
    \bibliography{main}
}

\clearpage
\newpage

\setcounter{page}{1}
\appendix
\onecolumn

\section{Proofs}
\subsection{Proof of Proposition~\ref{Prop: 1}}

 We aim to solve the optimization problem: \begin{equation*} \Gamma^* = \underset{\Gamma \in \mathbb{R}^{d \times d}}{\arg\!\min} ; \frac{1}{n} \left\Vert \Phi_I^\top - \Gamma \Phi_T^\top \right\Vert_F^2, \end{equation*} where $\Phi_I, \Phi_T \in \mathbb{R}^{d \times n}$ are given matrices, and $\left\Vert \cdot \right\Vert_F$ denotes the Frobenius norm.

To find the optimal $\Gamma^*$, we begin by expanding the objective function. Recall that the squared Frobenius norm of a matrix $A$ is given by $\left\Vert A \right\Vert_F^2 = \operatorname{Tr}(A^\top A)$. Therefore, we have:
\begin{align*}
    f(\Gamma) &= \frac{1}{n} \left\Vert \Phi_I^\top - \Gamma \Phi_T^\top \right\Vert_F^2 \\
    &= \frac{1}{n} \operatorname{Tr} \left[ \left( \Phi_I^\top - \Gamma \Phi_T^\top \right)^\top \left( \Phi_I^\top - \Gamma \Phi_T^\top \right) \right] \\
    &= \frac{1}{n} \operatorname{Tr} \left[ \Phi_I \Phi_I^\top - \Gamma \Phi_I \Phi_T^\top - \Phi_T \Phi_I^\top \Gamma^\top + \Gamma \Phi_T \Phi_T^\top \Gamma^\top \right].
\end{align*}

Let us define the covariance matrices:
\begin{align*}
    C_{II} &= \Phi_I \Phi_I^\top \in \mathbb{R}^{d \times d}, \\
    C_{IT} &= \Phi_I \Phi_T^\top \in \mathbb{R}^{d \times d}, \\
    C_{TI} &= \Phi_T \Phi_I^\top = C_{IT}^\top \in \mathbb{R}^{d \times d}, \\
    C_{TT} &= \Phi_T \Phi_T^\top \in \mathbb{R}^{d \times d}.
\end{align*}

Substituting these definitions into $f(\Gamma)$, we obtain:
\begin{align*}
    f(\Gamma) &= \frac{1}{n} \operatorname{Tr} \left[ C_{II} - \Gamma C_{IT} - C_{TI} \Gamma^\top + \Gamma C_{TT} \Gamma^\top \right].
\end{align*}

To find the minimizer, we compute the Jacobian of $f(\Gamma)$ with respect to $\Gamma$. Using standard matrix derivative identities, we have:
\begin{align*}
    \mathrm{J}_\Gamma f(\Gamma) &= \frac{1}{n} \left( - C_{IT}^\top - C_{TI} + 2 \Gamma C_{TT} \right) \\
    &= \frac{1}{n} \left( - C_{IT}^\top - C_{IT}^\top + 2 \Gamma C_{TT} \right) \quad \text{(since $C_{TI} = C_{IT}^\top$)} \\
    &= \frac{1}{n} \left( -2 C_{IT}^\top + 2 \Gamma C_{TT} \right).
\end{align*}
We observe that by choosing $\Gamma^* = C_{TI} C_{TT}^{-1}$, we will have
\begin{align*}
    \mathrm{J}_\Gamma f(\Gamma^*) = -\frac{2}{n} C_{IT}^\top + \frac{2}{n} \Gamma^* C_{TT} = \mathbf{0}.
\end{align*}
Therefore, $\Gamma^*$ is a stationary point in the optimization problem with a convex objective function and hence is an optimal solution to the minimization task.

\subsection{Conditional Entropy Interpretation of Scendi Score}

As discussed in the main text, in Equation \eqref{Eq: Schur component decomposition}, both image component $\Lambda_I$ and  text component $\Lambda_T$ are PSD matrices with unit trace. Furthermore, we have 
\begin{equation*}
   C_{II} = \mathrm{Tr}(\Lambda_I)\cdot \frac{1}{\mathrm{Tr}(\Lambda_I)}\Lambda_I + \bigl(1-\mathrm{Tr}(\Lambda_I)\bigr)\cdot \frac{1}{\mathrm{Tr}(\Lambda_T)}\Lambda_T 
\end{equation*}
Next, we consider the spectral decomposition of matrix $C_{II} = \sum_{i=1}^d \lambda_i \mathbf{v}_i\mathbf{v}_i^\top$ given its non-negative eigenvalues $\lambda_1\ge \cdots\ge \lambda_d$ and  orthonormal eigenvectors $\mathbf{v}_1,\ldots ,\mathbf{v}_n$. Following the orthonormality of the eigenvectors, we have the following for every $j\in\{1,\ldots,d\}$: 
\begin{equation*}
   \lambda_j = \mathrm{Tr}(\Lambda_I)\cdot \frac{1}{\mathrm{Tr}(\Lambda_I)}\mathbf{v}_j^\top\Lambda_I\mathbf{v}_j^\top + \bigl(1-\mathrm{Tr}(\Lambda_I)\bigr)\cdot \frac{1}{\mathrm{Tr}(\Lambda_T)}\mathbf{v}_j^\top\Lambda_T\mathbf{v}_j 
\end{equation*}
Therefore, if we define the $\mathrm{Mode}$ random variable over $\{1,\ldots , d\}$ with probabilities $\lambda_1,\ldots ,\lambda_d$, its unconditional Shannon entropy will be $H(\mathrm{Mode})=\sum_{i=1}^d\lambda_i \log(1/\lambda_i)$. On the other hand, if an adversary has the side knowledge of the text it can correctly predict $\mathrm{Mode}=j$ with probability $\mathbf{v}_j^\top \Lambda_T \mathbf{v}_j$. If we define $Y_{\mathrm{adv}}$as the correct prediction of this adversary when the text can be correctly mapped to the mode variable and else we define $Y_{\mathrm{adv}}=e$ as the error, then the conditional entropy will be:
\begin{align*}  H(\mathrm{Mode}|Y_{\mathrm{adv}})&= P(Y_{\mathrm{adv}}= e) H(\mathrm{Mode}|Y_{\mathrm{adv}}=e) + P(Y_{\mathrm{adv}}\neq e)H(\mathrm{Mode}|Y_{\mathrm{adv}}\neq e) \\
    &= P(Y_{\mathrm{adv}}= e) H(\mathrm{Mode}|Y_{\mathrm{adv}}=e) + P(Y_{\mathrm{adv}}\neq e)\times 0 \\
     &= P(Y_{\mathrm{adv}}= e) H(\mathrm{Mode}|Y_{\mathrm{adv}}=e) \\
     &= \Bigl(\sum_{j=1}^d v_j^\top \Lambda_I v_j \Bigr) \sum_{j=1}^d \frac{v_j^\top \Lambda_I v_j}{\sum_{t=1}^d v_t^\top \Lambda_I v_t}\log \frac{\sum_{t=1}^d v_t^\top \Lambda_I v_t}{v_j^\top \Lambda_I v_j} \\
     &=  \sum_{j=1}^d \bigl({v_j^\top \Lambda_I v_j}\bigr)\log \frac{\sum_{t=1}^d v_t^\top \Lambda_I v_t}{v_j^\top \Lambda_I v_j}
\end{align*}
Note that $\sum_{t=1}^d v_t^\top \Lambda_I v_t = \sum_{t=1}^d\mathrm{Tr}(v_t^\top \Lambda_I v_t) = \mathrm{Tr}(\sum_{t=1}^d v_tv_t^\top \Lambda_I)=\mathrm{Tr}( \Lambda_I)$ which implies that  
\begin{align*}  H(\mathrm{Mode}|Y_{\mathrm{adv}})&=  \sum_{j=1}^d \bigl({v_j^\top \Lambda_I v_j}\bigr)\log \frac{\mathrm{Tr}( \Lambda_I)}{v_j^\top \Lambda_I v_j}\\
&= \log(\mathrm{Tr}( \Lambda_I))\Bigl(\sum_{j=1}^d \bigl({v_j^\top \Lambda_I v_j}\bigr)\Bigr) + \sum_{j=1}^d \bigl({v_j^\top \Lambda_I v_j}\bigr)\log \frac{1}{v_j^\top \Lambda_I v_j}\\
&= \log(\mathrm{Tr}( \Lambda_I))\mathrm{Tr}( \Lambda_I) + \sum_{j=1}^d \bigl({v_j^\top \Lambda_I v_j}\bigr)\log \frac{1}{v_j^\top \Lambda_I v_j}
\end{align*}
which assuming that $\Lambda_I$ and $C_{II}$ share the same eigenvectors will provide
\begin{align*}  H(\mathrm{Mode}|Y_{\mathrm{adv}})&= \log(\mathrm{Tr}( \Lambda_I))\mathrm{Tr}( \Lambda_I) + \sum_{j=1}^d \bigl(\lambda^{(\Lambda_I)}_j\bigr)\log \frac{1}{\lambda^{(\Lambda_I)}_j} \\
&=  \sum_{j=1}^d\lambda^{(\Lambda_I)}_j \log \frac{\mathrm{Tr}(\Lambda_I)}{\lambda^{(\Lambda_I)}_j}
\end{align*}
Note that the above provides our definition of the Schur-Complement-Entropy for the image part $ \mathrm{Scendi_I}$ and the text part $\mathrm{Scendi_T}$ as follows:
\begin{align}
    \mathrm{Scendi}_I(x_1,\ldots ,x_n)\, =& \,  \sum_{j=1}^d\lambda^{(\Lambda_I)}_j \log \frac{\mathrm{Tr}(\Lambda_I)}{\lambda^{(\Lambda_I)}_j} \\
    \mathrm{Scendi}_T(x_1,\ldots ,x_n)\, :=& \,  \sum_{j=1}^d\lambda^{(\Lambda_T)}_j \log \frac{\mathrm{Tr}(\Lambda_T)}{\lambda^{(\Lambda_T)}_j}
\end{align}

where $\lambda^{(\Lambda_I)}_j$ denotes the $j$th eigenvalue of matrix $\Lambda_I$ and $  \mathrm{Tr}(\Lambda_I)= \sum_{j=1}^d \lambda^{(\Lambda_I)}_j$ is the sum of the eigenvalues. Note that we follow the same definition for the text part $\Lambda_T$. 

\section{Limitations}
The Scendi framework is only compatible with cross‑modal embeddings, such as those produced by CLIP. When such embeddings are unavailable for a given data modality, evaluators cannot use Scendi to measure diversity. Extending Scendi to modalities without cross‑modal embeddings remains an open challenge and a promising direction for future work.

\section{Additional Individual Image Decomposition Results via SC-Based Method}
In this section, we present additional CLIP decomposition results for randomly selected pairs of ImageNet labels. The correction matrix was computed using the captioned MSCOCO dataset. The experimental setup follows the approach illustrated in Figure \ref{fig:ostrich and pumpkin interpretability}. We generated images containing predominantly two concepts and applied the SC-based method for decomposition. Subsequently, we measured the cosine similarity between the corrected and regular CLIP embeddings and the CLIP-embedded ImageNet samples. The top four images with the highest similarity scores are reported. These results demonstrate the effectiveness of the Schur Complement method in decomposing directions present in generated images.

\begin{figure*}
    \centering
    \includegraphics[width=0.95\linewidth]{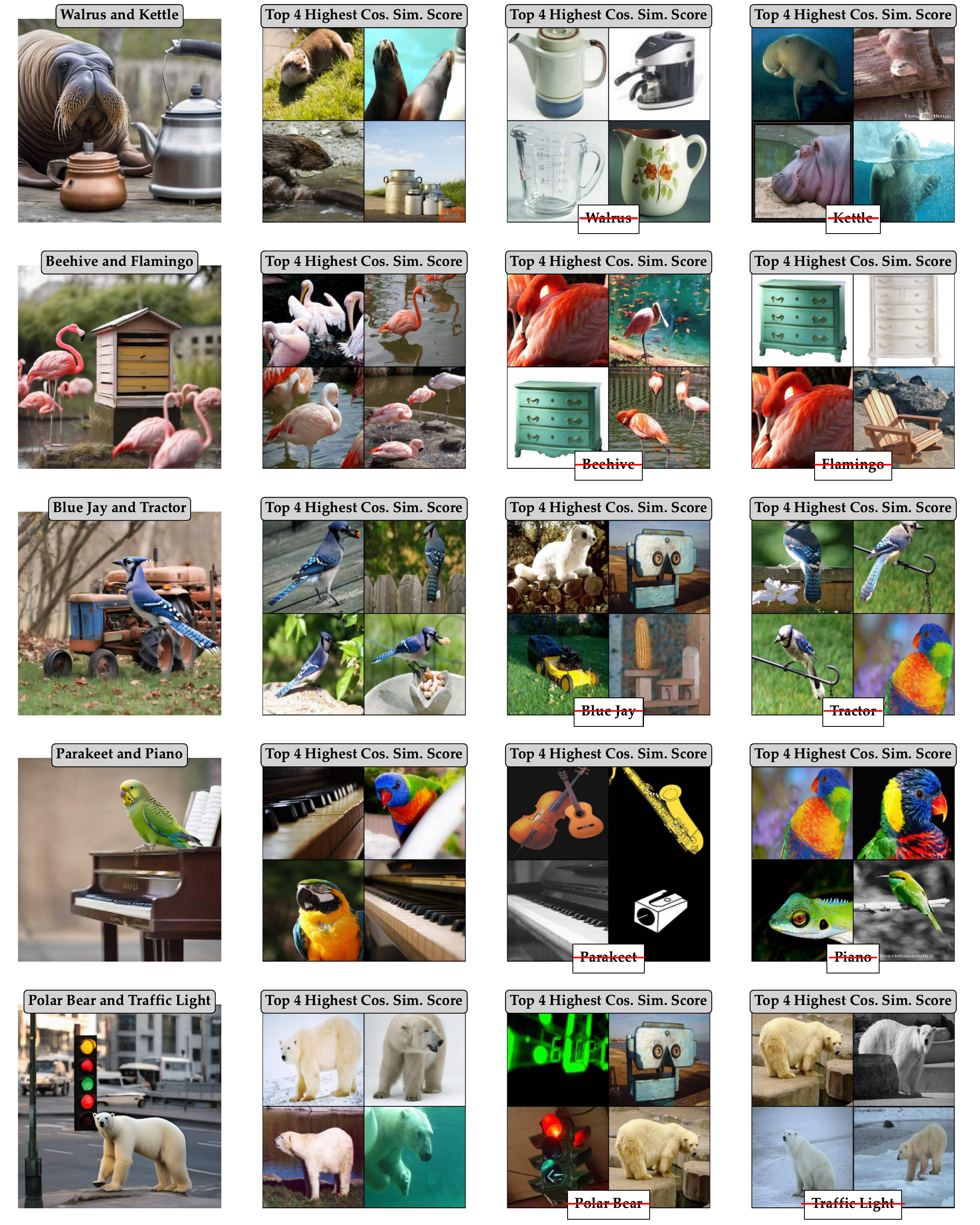}
    \caption{Diagram presenting the decomposition of SDXL generated images of two random labels from ImageNet. First column presents the generated image of a pair. Second column presents four images from ImageNet with highest Cosine Similarity Score. Third and Fourth columns showcase feature removal from the image.}
    \label{fig:imagenet sdxl clipscore batch 1}
\end{figure*}

\begin{figure*}
    \centering
    \includegraphics[width=0.95\linewidth]{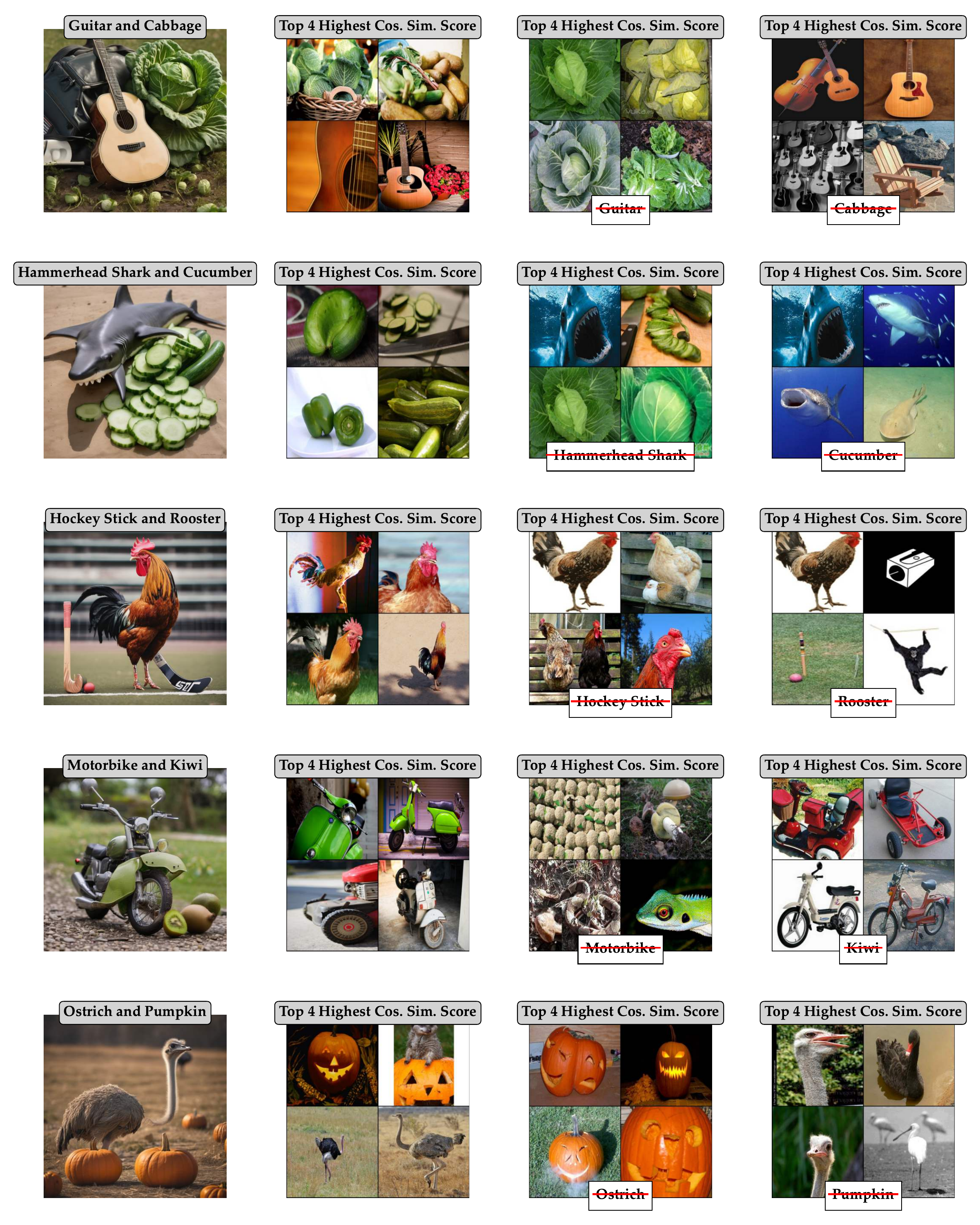}
    \caption{Diagram presenting the decomposition of SDXL generated images of two random labels from ImageNet. First column presents the generated image of a pair. Second column presents four images from ImageNet with highest Cosine Similarity Score. Third and Fourth columns showcase feature removal from the image.}
    \label{fig:imagenet sdxl clipscore batch 2}
\end{figure*}

\begin{figure*}
    \centering
    \includegraphics[width=0.95\linewidth]{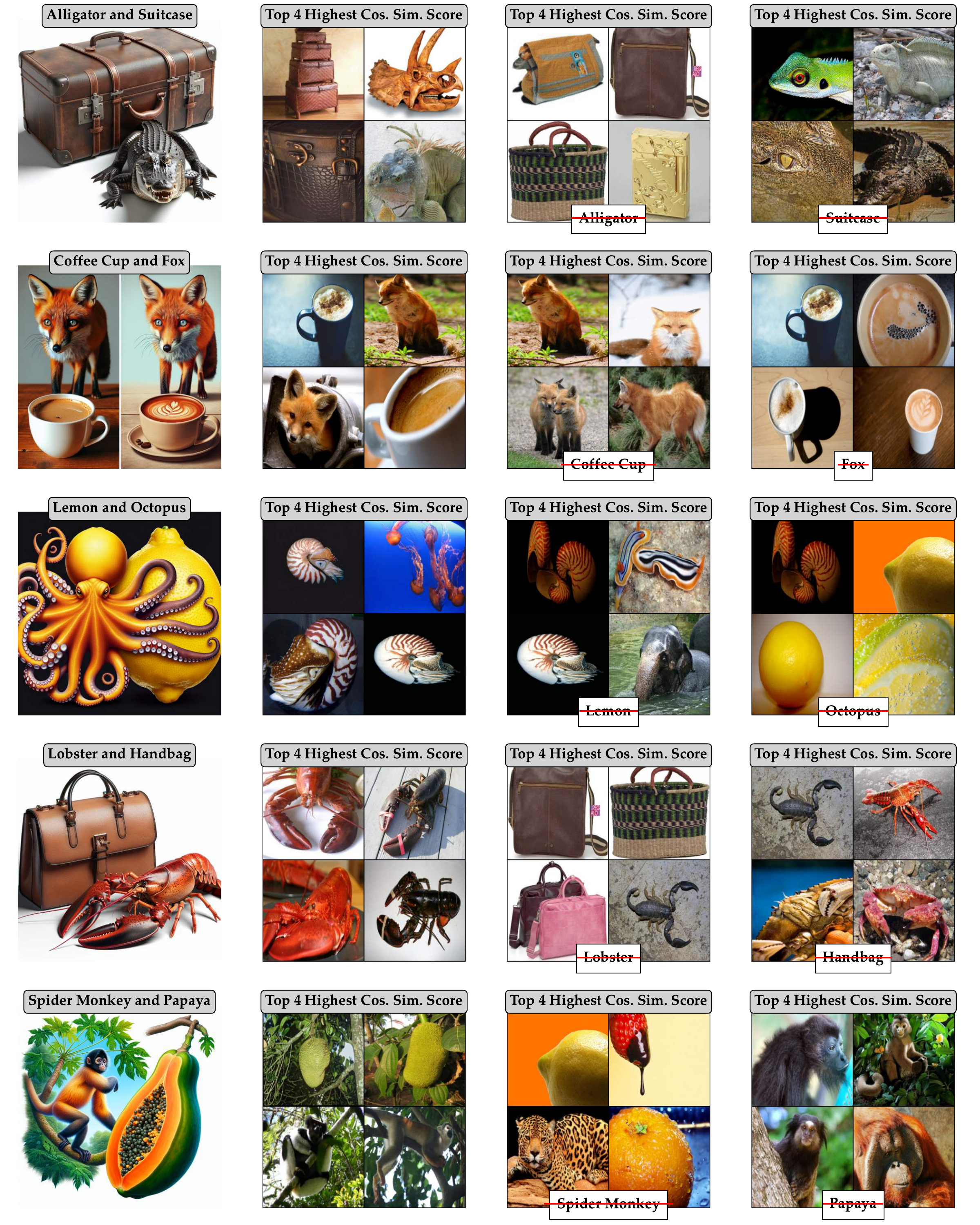}
    \caption{Diagram presenting the decomposition of DALL-E 3 generated images of two random labels from ImageNet. First column presents the generated image of a pair. Second column presents four images from ImageNet with highest Cosine Similarity Score. Third and Fourth columns showcase feature removal from the image.}
    \label{fig:imagenet dalle3 clipscore batch 1}
\end{figure*}

\begin{figure*}
    \centering
    \includegraphics[width=0.95\linewidth]{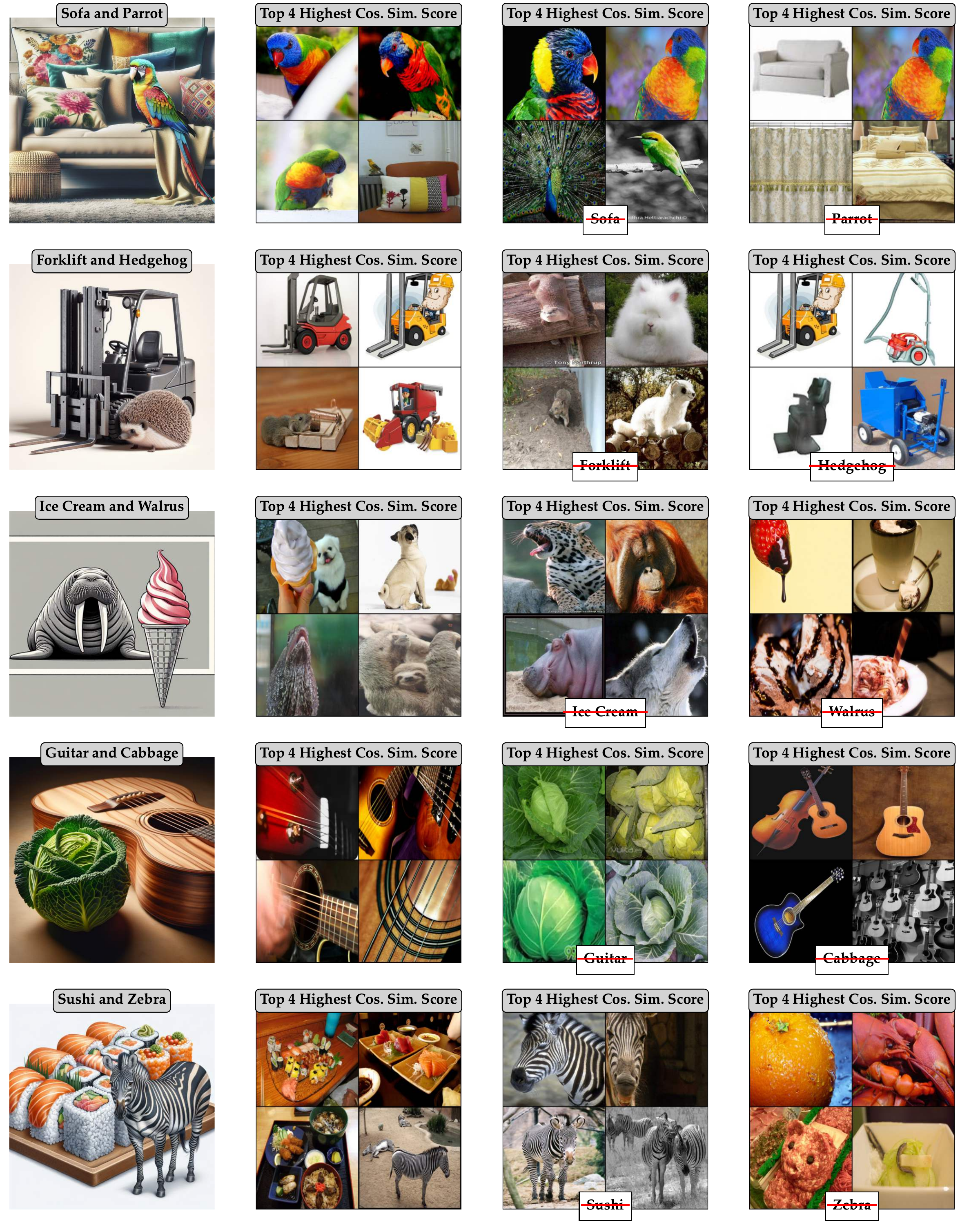}
    \caption{Diagram presenting the decomposition of DALL-E 3 generated images of two random labels from ImageNet. First column presents the generated image of a pair. Second column presents four images from ImageNet with highest Cosine Similarity Score. Third and Fourth columns showcase feature removal from the image.}
    \label{fig:imagenet dalle3 clipscore batch 2}
\end{figure*}

Results for synthetic images generated using SDXL are shown in Figures \ref{fig:imagenet sdxl clipscore batch 1} and \ref{fig:imagenet sdxl clipscore batch 2}. Corresponding results for DALL-E 3 are presented in Figures \ref{fig:imagenet dalle3 clipscore batch 1} and \ref{fig:imagenet dalle3 clipscore batch 2}. Notably, the Schur Complement-based decomposition successfully isolates and removes image directions corresponding to a text condition that describes the concept to be excluded.

\begin{figure*}
    \centering
    \includegraphics[width=0.95\linewidth]{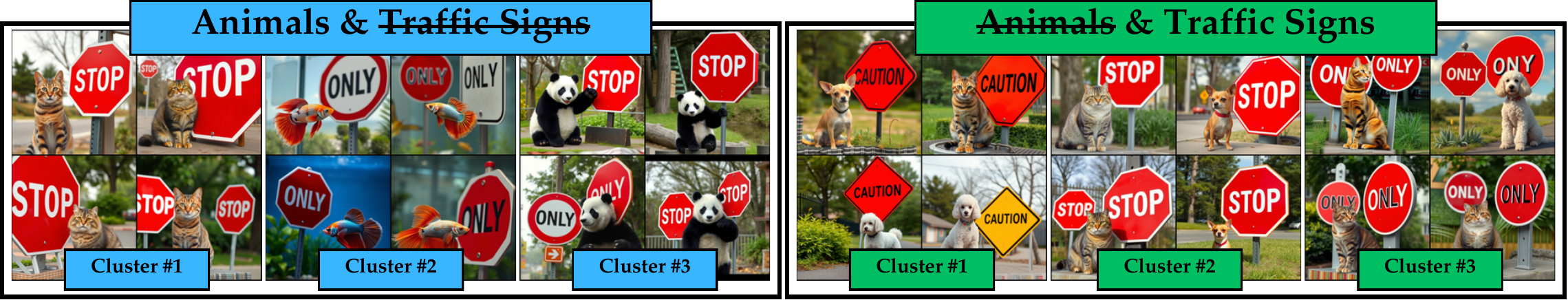}
    \caption{Identified Kernel PCA clusters on the synthetic dataset composed of random animals with traffic signs.}
    \label{fig:animals signs clusters}
\end{figure*}

To expand on the results in Figure \ref{fig:animals fruits dalle3}, we constructed a dataset of animals with traffic signs using FLUX.1-schnell \cite{flux_2024}. Figure \ref{fig:animals signs clusters} illustrates that after canceling either of the subjects using the Schur complement method, Kernel-PCA clusters according to the remaining concepts in the image. 

\begin{figure*}
    \centering
    \includegraphics[width=\linewidth]{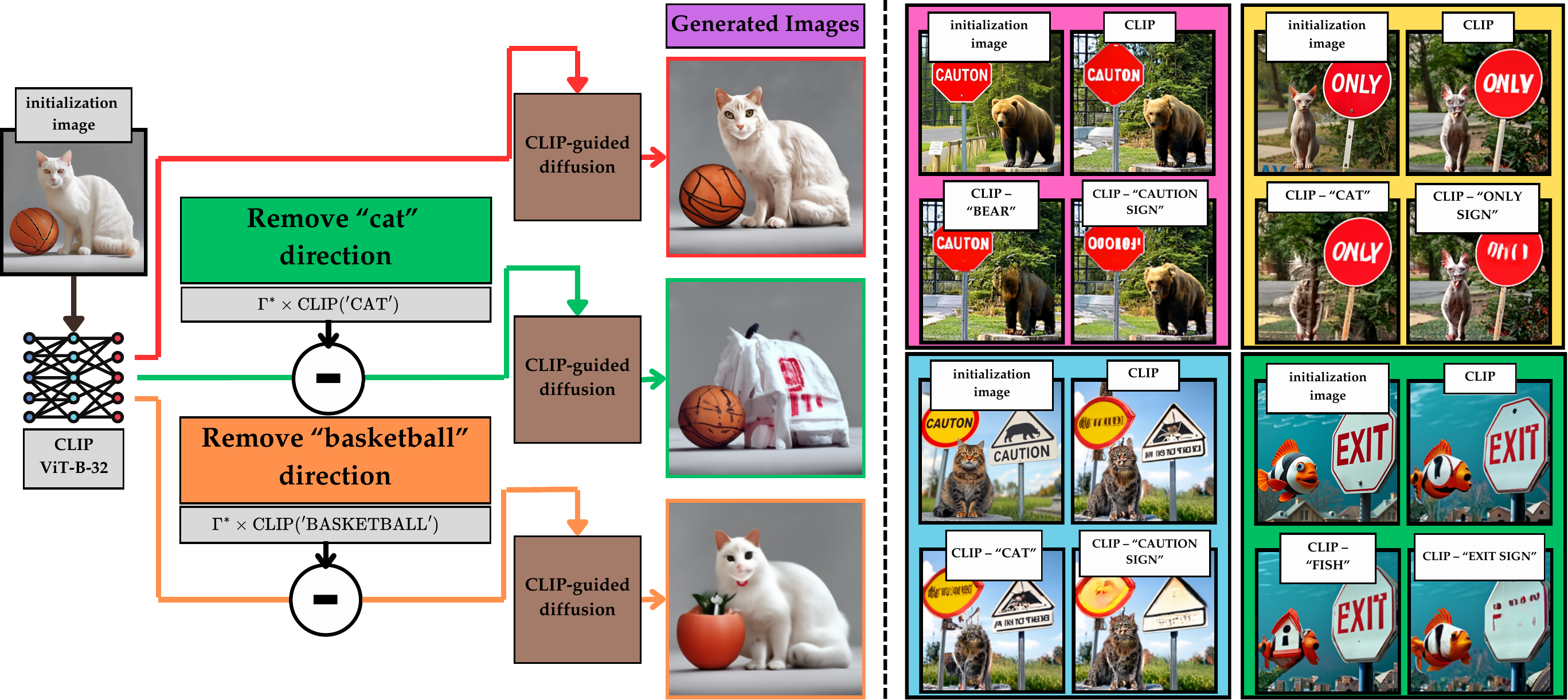}
    \caption{CLIP-guided diffusion process. Starting from an 'initialization image,' generation is guided by CLIP embeddings. The baseline (red arrow) shows unchanged denoising. Adjusted CLIP-guided results (green and orange arrows) show denoised images after removing one of the subjects. Additional clip-guided denoised samples are shown on the right.}
    \label{fig:diffusion clip}
\end{figure*}

Moreover, to test how CLIP correction affects the underlying directions of concepts, we applied the CLIPDiffusion \cite{Kim_2022_CVPR} framework to edit the image according to different CLIP embeddings. Figure \ref{fig:diffusion clip} illustrates the setup of the problem, where we edit the 'initialization image' that consists of two subjects: a cat and a basketball. We then denoise and guide the generation according to three different embeddings: the unchanged CLIP embedding of the 'initialization image', the modified CLIP by a 'cat' direction, or the modified CLIP by a 'basketball' direction. We show that after removing a concept direction, the denoiser is no longer rewarded for generating the corresponding concept, which is reflected in the denoised images. After correction, the basketball resembles a bowl with plants, and the cat loses its features. We also note that in both cases, the other object remains intact. To further showcase these results, we performed the same diffusion on the animals with traffic signs dataset, shown on the side of Figure \ref{fig:diffusion clip}.

\section{Additional Results on Diversity Evaluation}
\begin{figure*}
    \centering
    \includegraphics[width=0.9\linewidth]{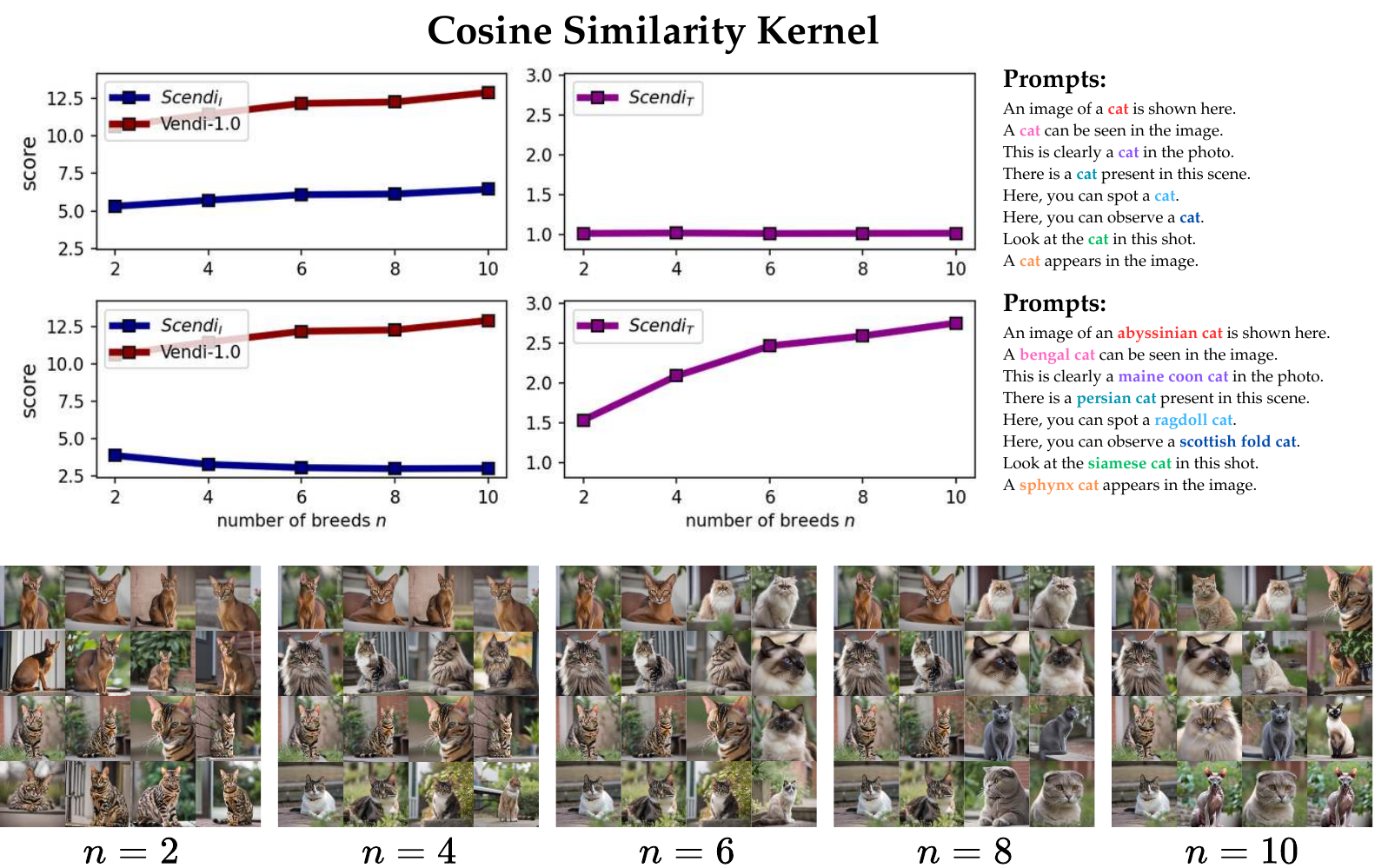}
    \caption{Plots by cancelling out 'cat' and specific cat breed prompts (Cosine Similarity Kernel)}
    \label{fig:cosine cat breed diversity}
\end{figure*}

\begin{figure*}
    \centering
    \includegraphics[width=\linewidth]{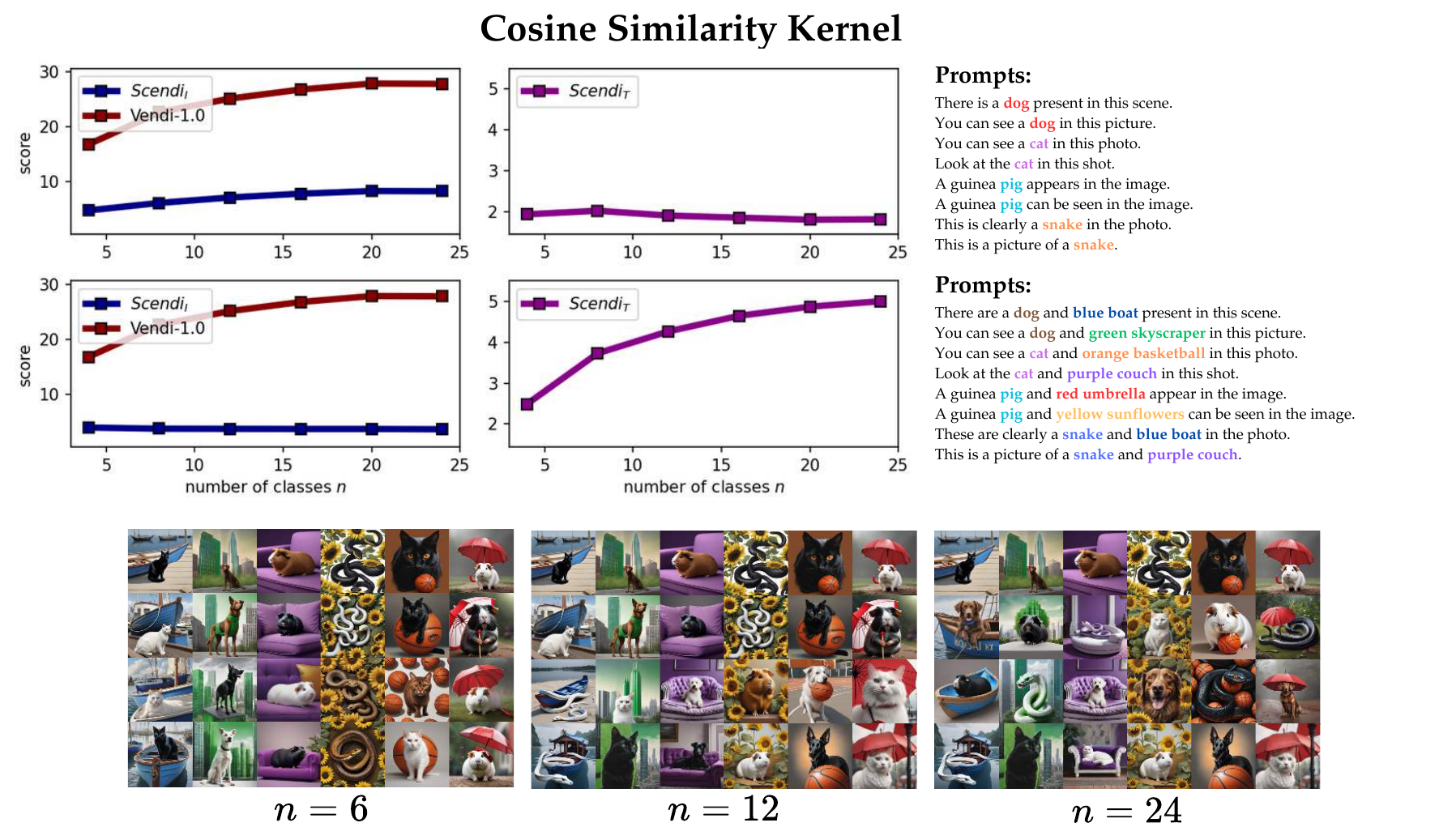}
    \caption{Plots by cancelling out animal name and specific object types prompts (Cosine Sim Kernel)}
    \label{fig:cosine animals with objects diversity}
\end{figure*}

To further validate the findings presented in the main text, we conducted a similar experiment (Figure \ref{fig:cat breed gaussian diversity plots}) using the Cosine Similarity Kernel. The results confirm that the diversity trends observed in the main text persist under a finite-dimensional kernel. Figures \ref{fig:cosine cat breed diversity} and \ref{fig:cosine animals with objects diversity} illustrate the variation in Scendi diversity when conditioned on different text prompts.

\begin{figure*}
    \centering
    \includegraphics[width=\linewidth]{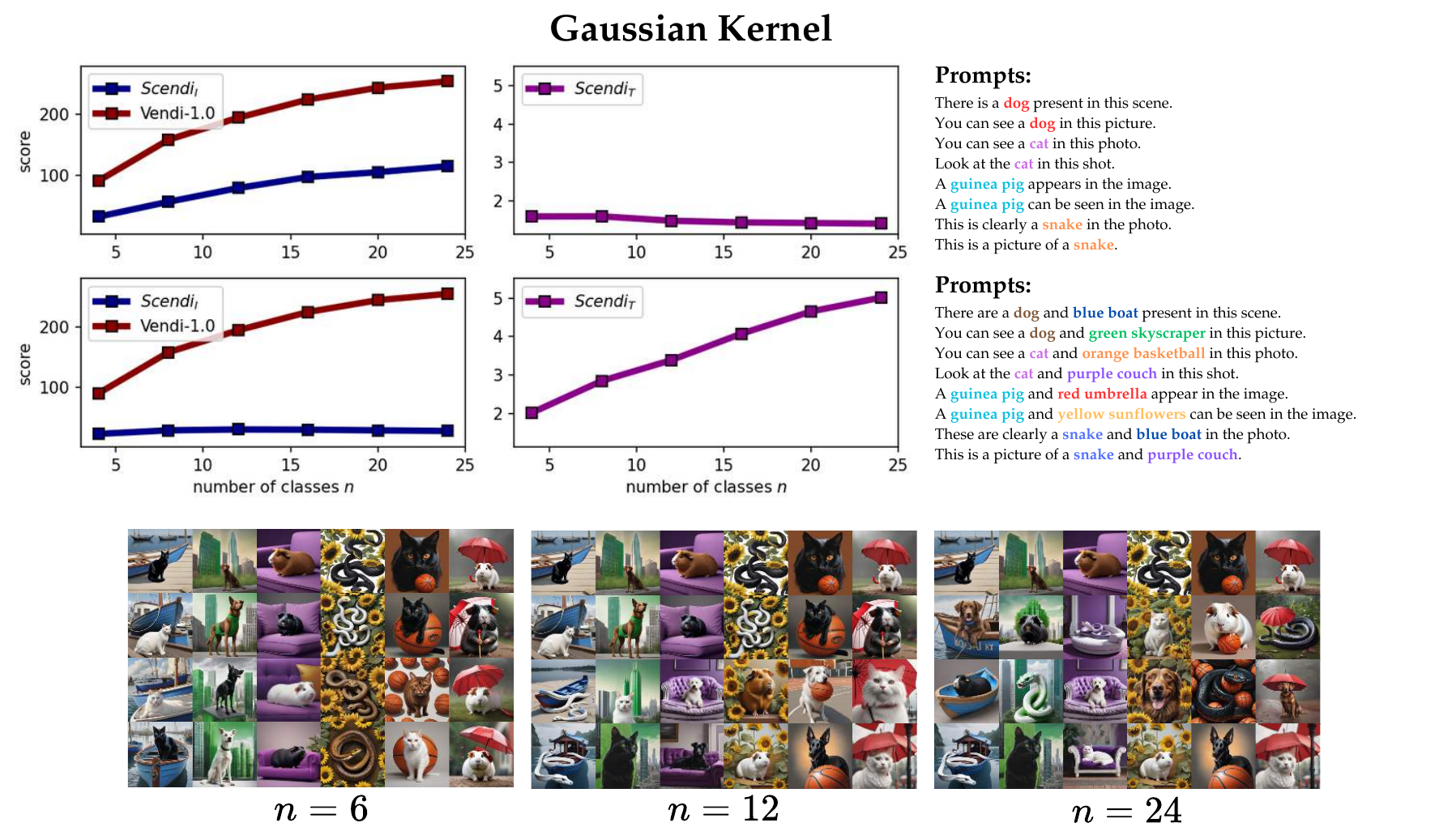}
    \caption{Evaluated Scendi and Vendi scores with Gaussian Kernel on different animals with objects.}
    \label{fig:gaussian animals with objects plot}
\end{figure*}

Similar to Figure~\ref{fig:cat breed gaussian diversity plots}, we conducted similar experiment with animals and objects dataset in Figure~\ref{fig:gaussian animals with objects plot}. Our resuts mirror previous findings, strengthening the proposed diversity evaluation metric in measuring subject quantity related diversity. 

\begin{figure*}[h]
    \centering
    \includegraphics[width=\linewidth]{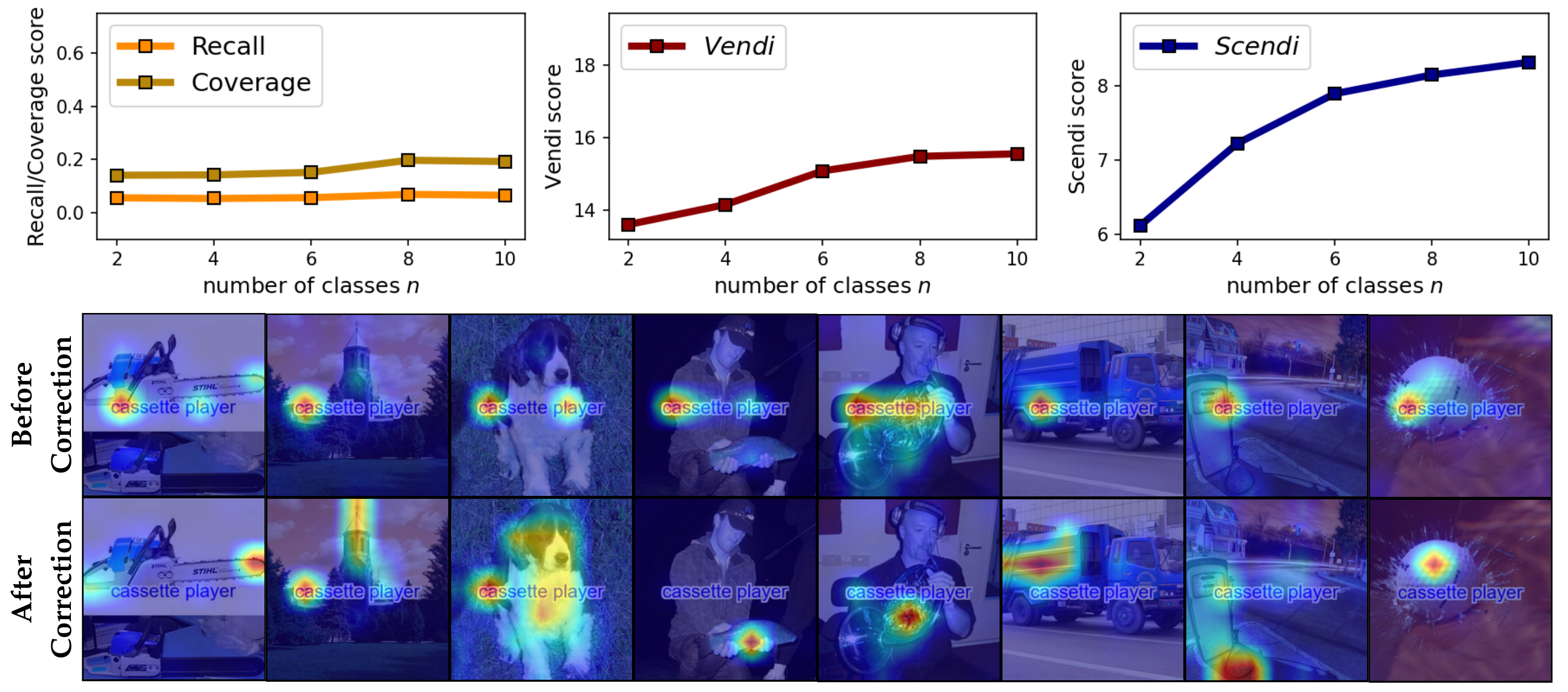}
    \caption{Evaluated Scendi, Vendi, Recall and Coverage scores with Gaussian Kernel on ImageNet with overlayed text.}
    \label{fig:gaussian imagenet caption diversity salience maps}
\end{figure*}

Moreover, we evaluate the diversity of typographically attacked ImageNet samples in Figure~\ref{fig:gaussian imagenet caption diversity salience maps}. Specifically, we overlay the text "cassette player" onto images from 10 different ImageNet classes and measure diversity as the number of distinct classes increases. The presence of overlaid text diverts CLIP’s sensitivity away from image content, causing it to encode the direction indicated by the text instead.

To illustrate this effect, we visualize salience maps of CLIP embeddings given a prompt referring to an object behind the overlayed text. The results show that CLIP is highly sensitive to centrally placed text, even when it is unrelated to the prompted object. However, applying SC-based decomposition mitigates this bias. This correction is reflected in the diversity plots, where $Scendi_I$ increases rapidly as the number of ImageNet classes grows, whereas Vendi, Coverage, and Recall exhibit much weaker correlations with class diversity.

\begin{figure*}
    \centering
     \includegraphics[width=0.93\linewidth]{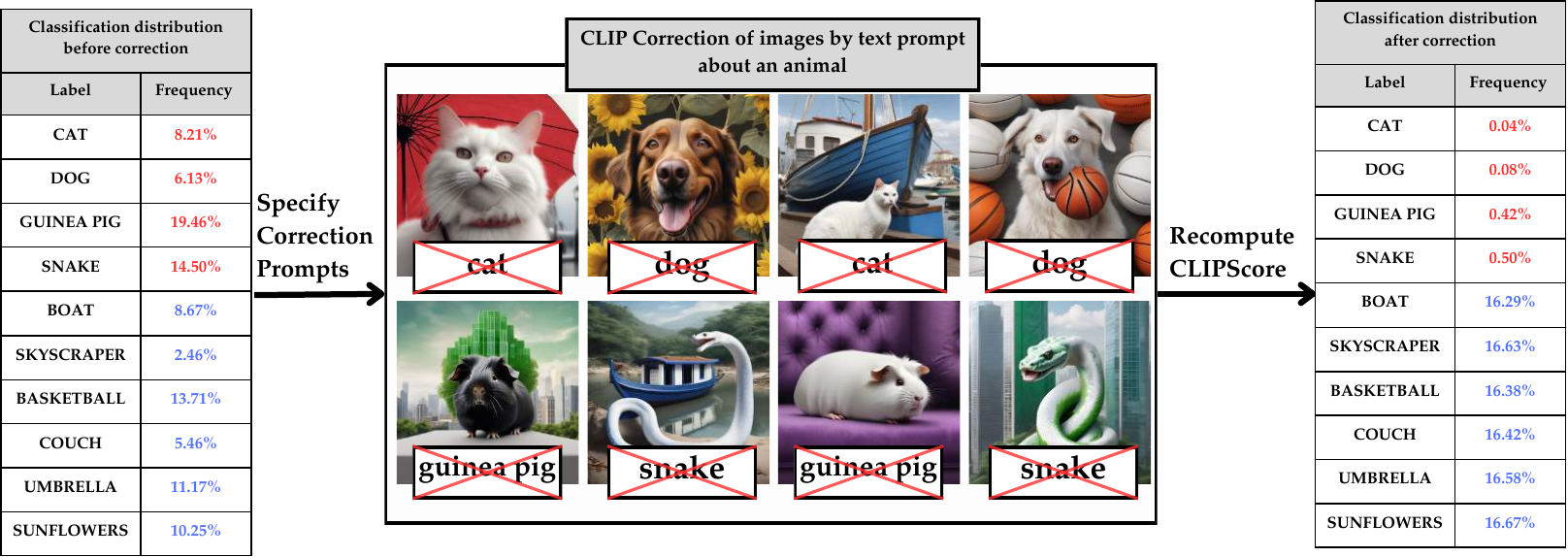}
    \caption{Classification distribution before and after CLIP correction on SDXL~\cite{podell2024sdxl} generated images of animals with objects in the background}
    \label{fig:clipscore animals objects t2i sdxl}
    \vspace{-5mm}
\end{figure*}

\begin{figure*}[h]
    \centering
    \begin{subfigure}[b]{\linewidth}
        \includegraphics[width=\linewidth]{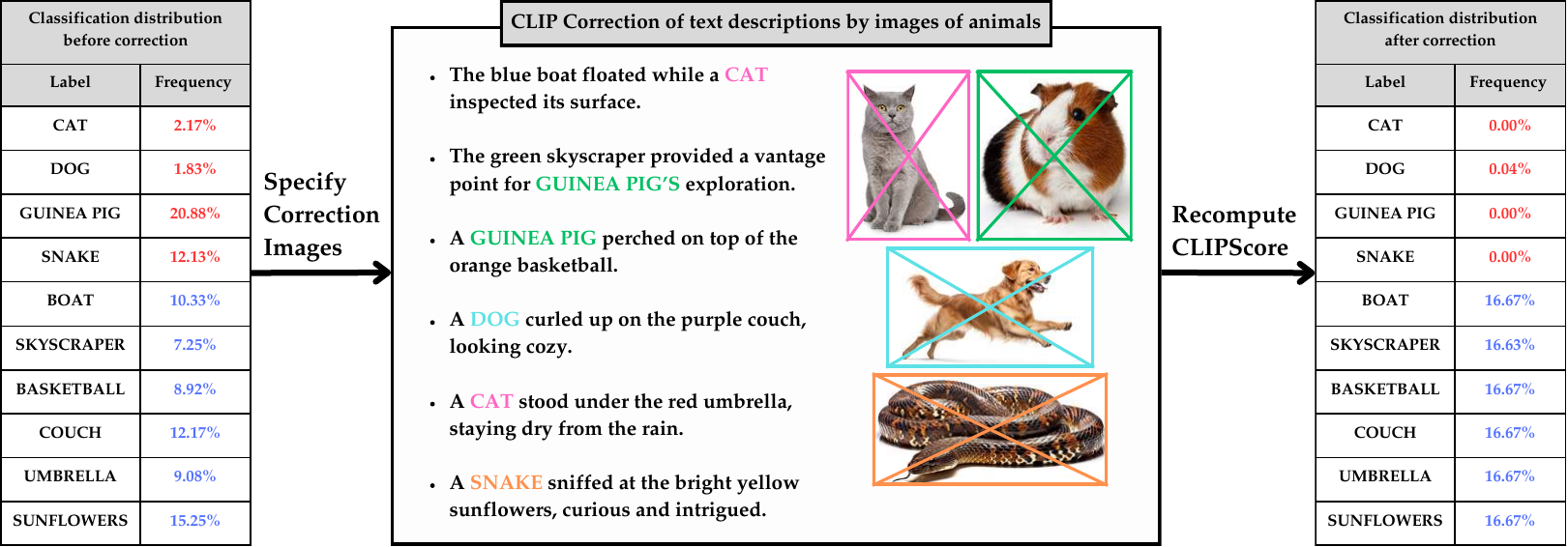}
        \caption{Effect of cancelling 'animals' direction from text given images of animals}
    \end{subfigure}
    \hfill
    \begin{subfigure}[b]{\linewidth}
        \includegraphics[width=\linewidth]{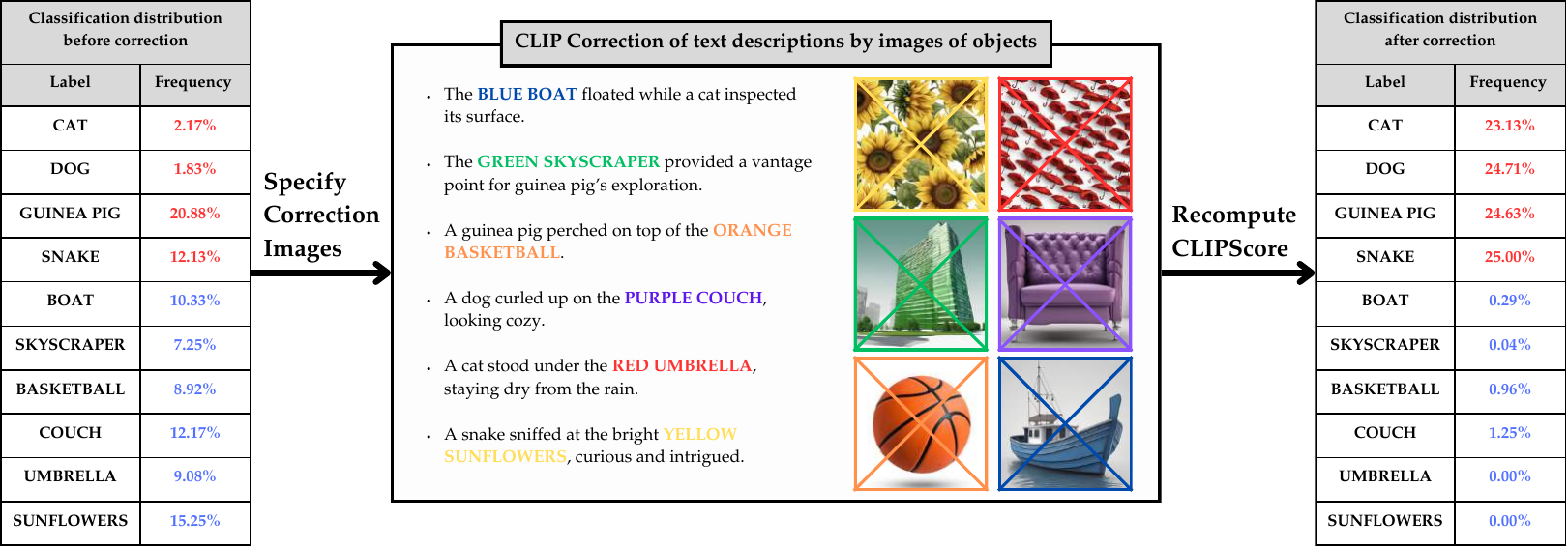}
        \caption{Effect of cancelling 'objects' direction from text given images of objects}
    \end{subfigure}
    \caption{Evaluating the CLIPScore on GPT-4o generated captions of animals with objects}
    \label{fig:clipscore animals objects i2t gpt4o both}
\end{figure*}

\section{Additional Experiments on the Image Captioning Task}  
In the main text, we discussed SC-based decomposition for text-to-image models and demonstrated how images can be decomposed given text prompts. Here, we show that the reverse process is also possible. Specifically, we explored decomposing captions based on their corresponding images.  

The experimental setup mirrors that of Figure \ref{fig:clipscore animals objects t2i sdxl}, with the key difference being that, instead of generating images for text prompts, we generated captions for the corresponding images. For this task, we used \emph{gpt4o-mini} as the captioning model. Figure \ref{fig:clipscore animals objects i2t gpt4o both} illustrates the experimental setup.  

We selected images closely aligned with the concept we aimed to remove. For instance, to eliminate the "cat" direction in text, we used an image of a cat against a white background to better isolate the concept. After applying corrections for "animals" or "objects" in the text prompt, we observed successful decomposition, as reflected in the second column: the corrected CLIP embedding is no longer sensitive to the removed concept.  

These findings highlight the versatility of the Scendi method, demonstrating its applicability across a wide range of tasks that rely on a shared embedding space.

\section{Robustness of Scendi}
\begin{figure}
    \centering
    \includegraphics[width=\linewidth]{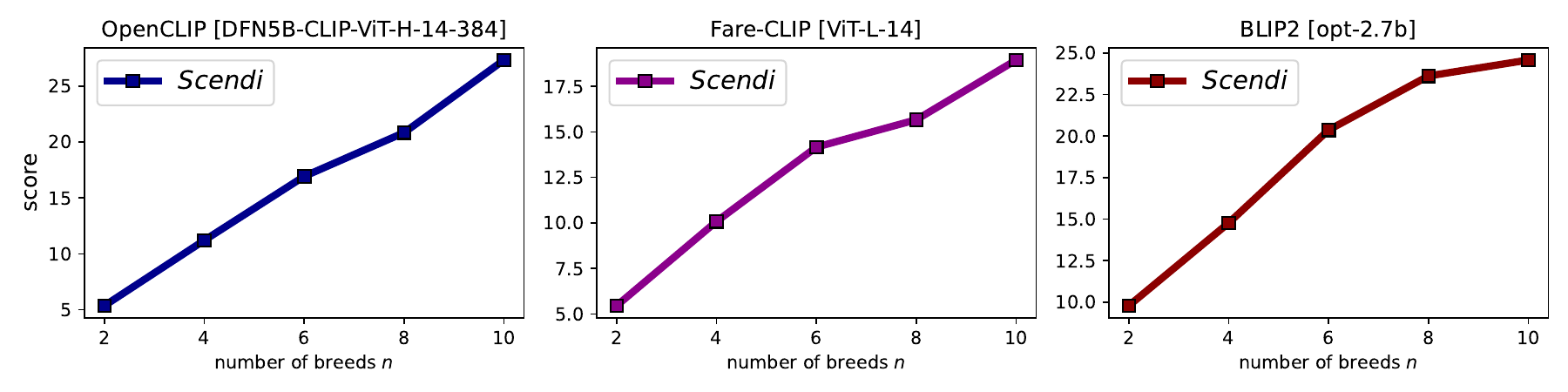}
    \caption{Figure 5's Scendi evaluation of different embeddings:  OpenCLIP (left), Robust FAIR-CLIP (middle), BLIP2 (right)}
    \vspace{-4mm}
    \label{fig:better-clip}
\end{figure}

We note that the robustness of the Scendi framework depends on the choice of underlying embedding. To address limitations in CLIP, several alternatives have been introduced, such as FairCLIP~\cite{luo2024fairclipharnessingfairnessvisionlanguage}. Because Scendi is compatible with any cross‑modal embedding, we evaluated diversity using three additional variants: OpenCLIP~\cite{ilharco_gabriel_2021_5143773}, FairCLIP, and BLIP2~\cite{li2022blip} on the SDXL generated cat breed dataset. The results appear in Figure~\ref{fig:better-clip}. We observe that Scendi preserves the qualities of a diversity metric that increases as we introduce more breeds into the data pool.

\section{Results with Naive Text Embedding Subtraction without considering the adjustment matrix $\Gamma^*$}
In the given task setting, it may seem intuitive to assume that the difference between $\Phi_I$ and $\Phi_T$ would yield a similar outcome as the SC-based decomposition. To test this hypothesis, we compared the SC-based decomposition with a "naive" method, defined as $\Phi_I - \Phi_T$. In this naive approach, the learned correction matrix $\Gamma^*$ is replaced with an identity matrix, effectively omitting its computation. Our experiments reveal that such a decomposition usually fails to achieve the desired results and often leads to a loss of coherent directionality in the embedding space.

\begin{figure*}[h]
    \centering
    \begin{subfigure}[b]{0.8\linewidth}
        \includegraphics[width=\linewidth]{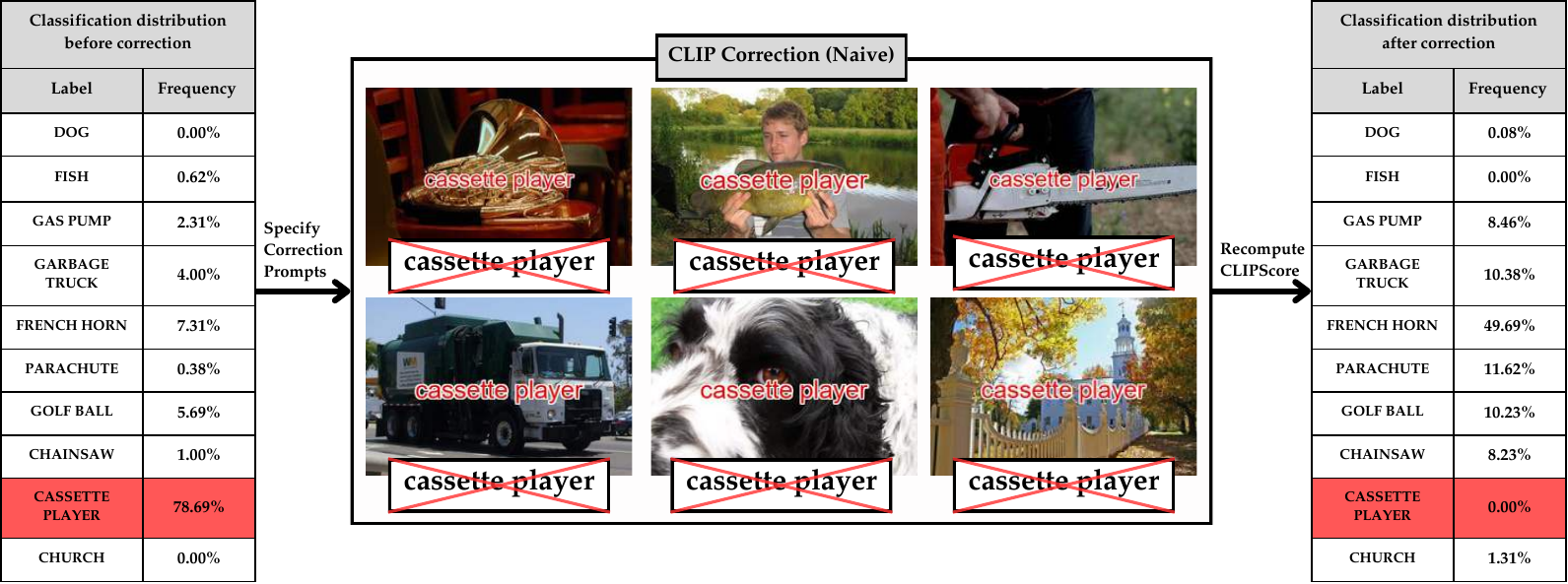}
    \end{subfigure}
    \hfill
    \begin{subfigure}[b]{0.8\linewidth}
        \includegraphics[width=\linewidth]{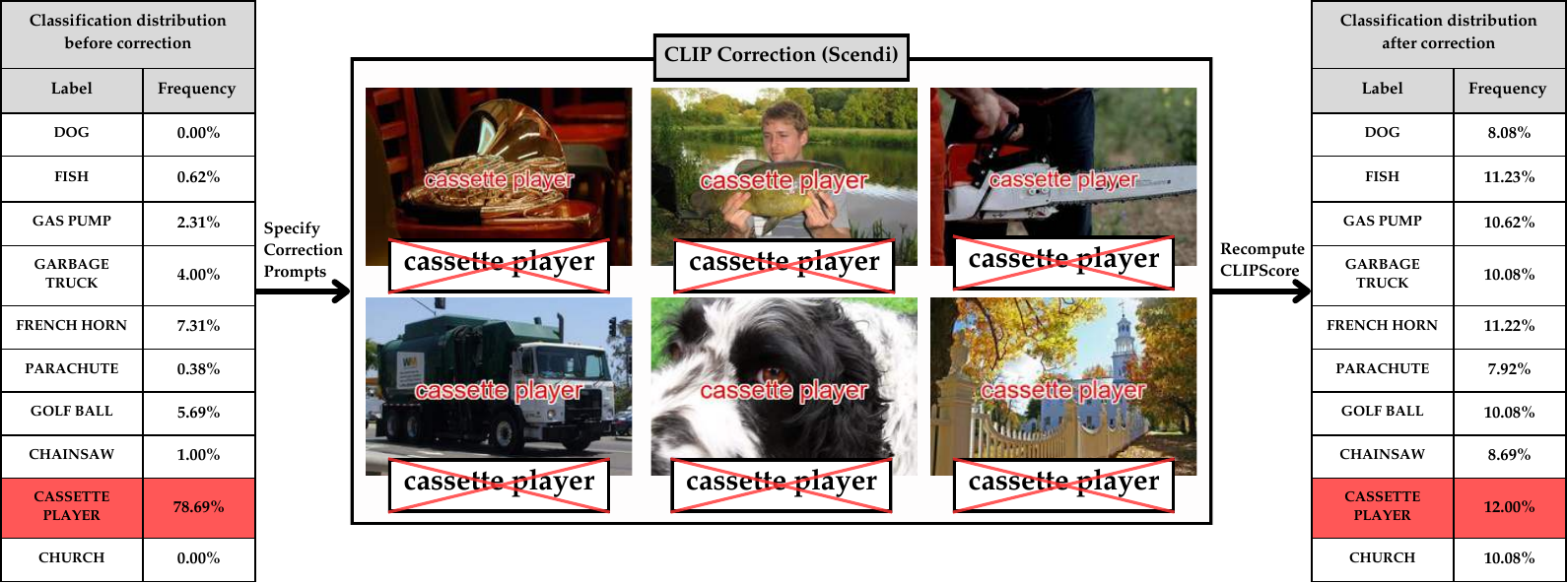}
    \end{subfigure}
    \hfill
    \caption{Effect of removing encoded "cassette player" text on top of ImageNet samples. Top figure represents naive method and bottom figure represents SC method.}
    \label{fig:caption clipscore with naive}
\end{figure*}

To evaluate the performance of the naive embedding subtraction method, we used the typographic attack dataset, which consists of 10 ImageNet classes where misleading text is overlaid on the images. We measured classification accuracy before and after decomposition. Figure \ref{fig:caption clipscore with naive} shows the distribution of classifications for images engraved with the text "cassette player." CLIP's classification is heavily biased towards "cassette player," despite the underlying images belonging to a different class. After decomposition, the naive method removes the direction corresponding to the text but results in a skew towards "french horn," even though the image distribution is uniform across all 10 classes. In contrast, the SC-based decomposition corrects the embeddings, making them sensitive to the underlying images rather than the engraved text.

\begin{figure*}[h]
    \centering
    \begin{subfigure}[b]{0.95\linewidth}
        \includegraphics[width=\linewidth]{figures/clustering/imagenette_captions/captions_clusters_before.pdf}
    \end{subfigure}
    \hfill
    \begin{subfigure}[b]{0.95\linewidth}
        \includegraphics[width=\linewidth]{figures/clustering/imagenette_captions/captions_clusters_after.pdf}
    \end{subfigure}
    \hfill
    \begin{subfigure}[b]{0.95\linewidth}
        \includegraphics[width=\linewidth]{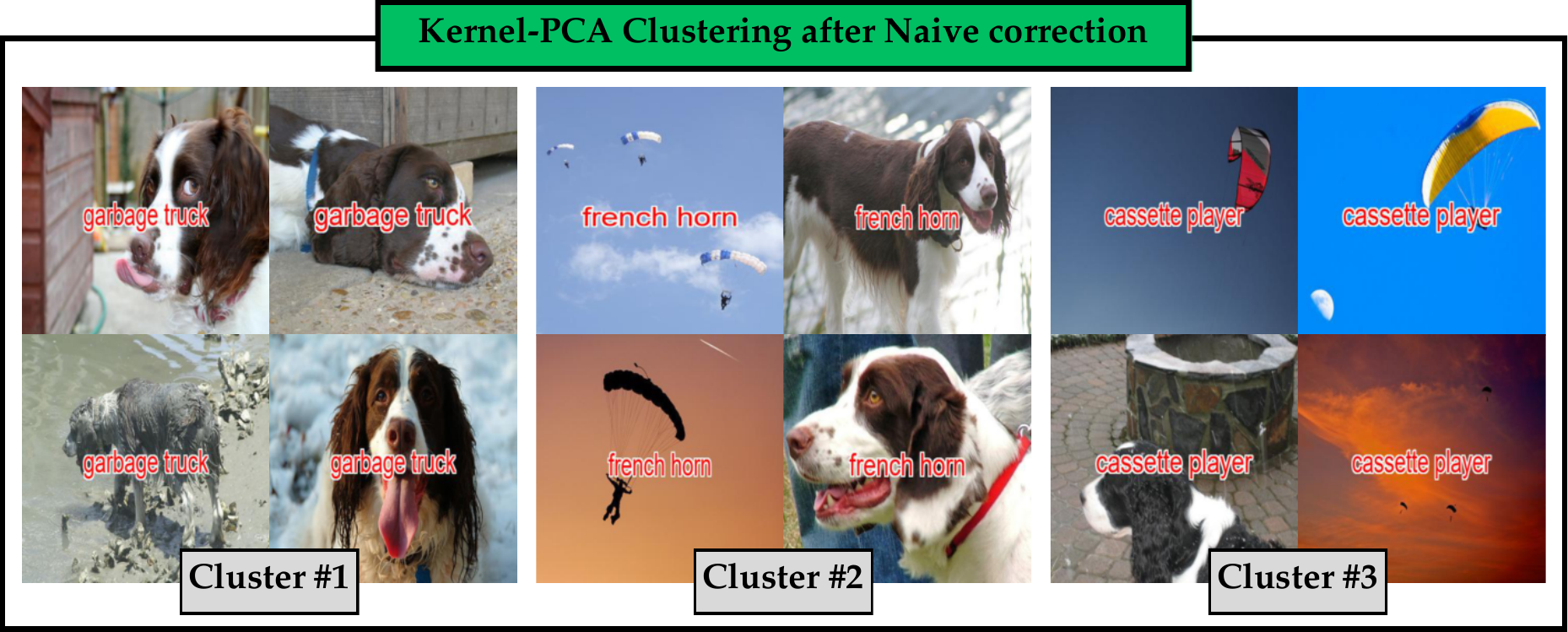}
    \end{subfigure}
    \hfill
    \caption{Kernel PCA clusters before and after CLIP correction on the captioned ImageNet dataset, comparing the Scendi decomposition approach with the naive approach.}
    \label{fig:caption clusters with naive}
\end{figure*}

To further demonstrate the effectiveness of the SC method, we compared kernel PCA clusters in Figure \ref{fig:caption clusters with naive}. The clustering results for the naive decomposition closely resemble those without any correction, indicating that this method does not address the typographic attack. On the other hand, SC-based decomposition significantly improves the clustering by accurately resolving the misleading text directionality.

\begin{figure*}
    \centering
    \includegraphics[width=\linewidth]{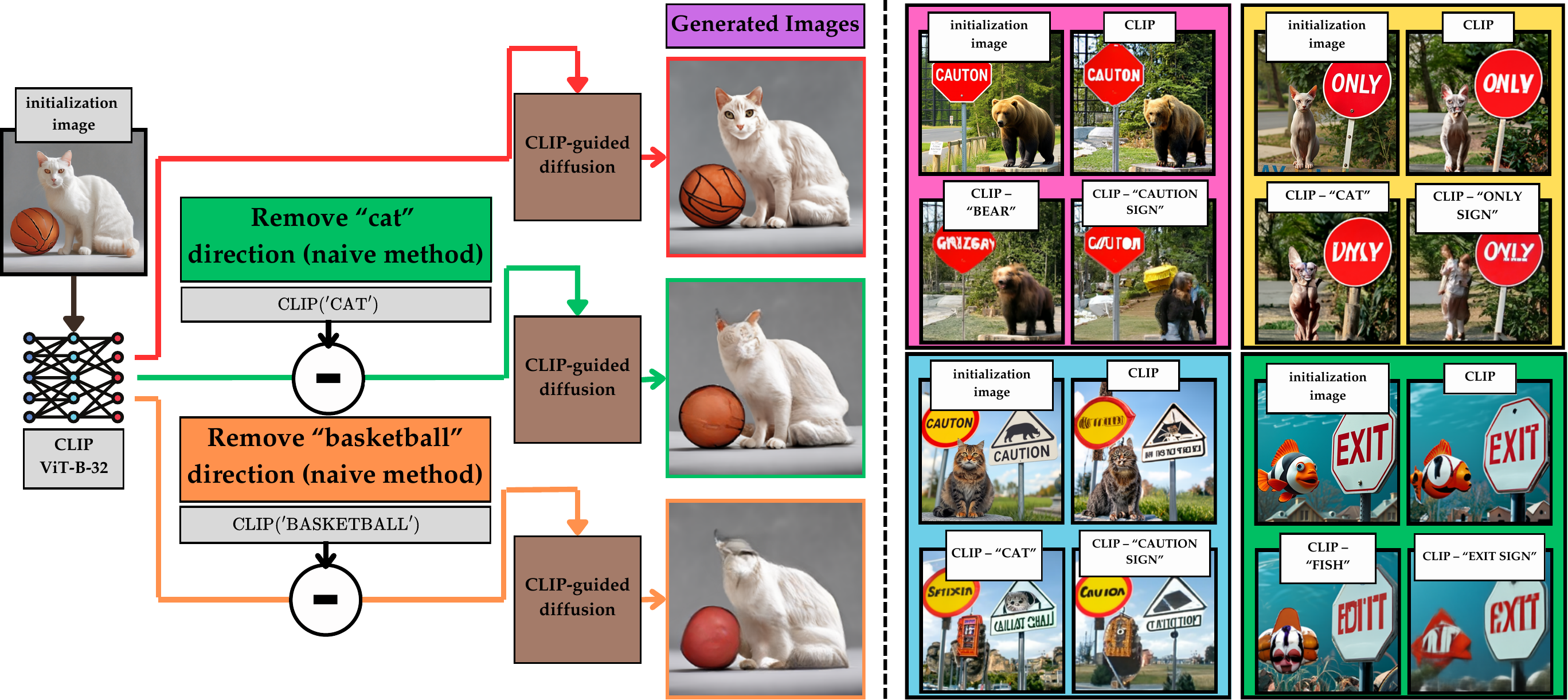}
    \caption{CLIP-guided diffusion process with Naive text embedding cancellation. Starting from an 'initialization image,' generation is guided by CLIP embeddings. The baseline (red arrow) shows unchanged denoising. Naive adjusted CLIP-guided results (green and orange arrows) show  image embeddings after subtracting the text CLIP embedding.}
    \label{fig:diffusion clip naive diagram}
\end{figure*}

\begin{figure*}
    \centering
    \includegraphics[width=0.8\linewidth]{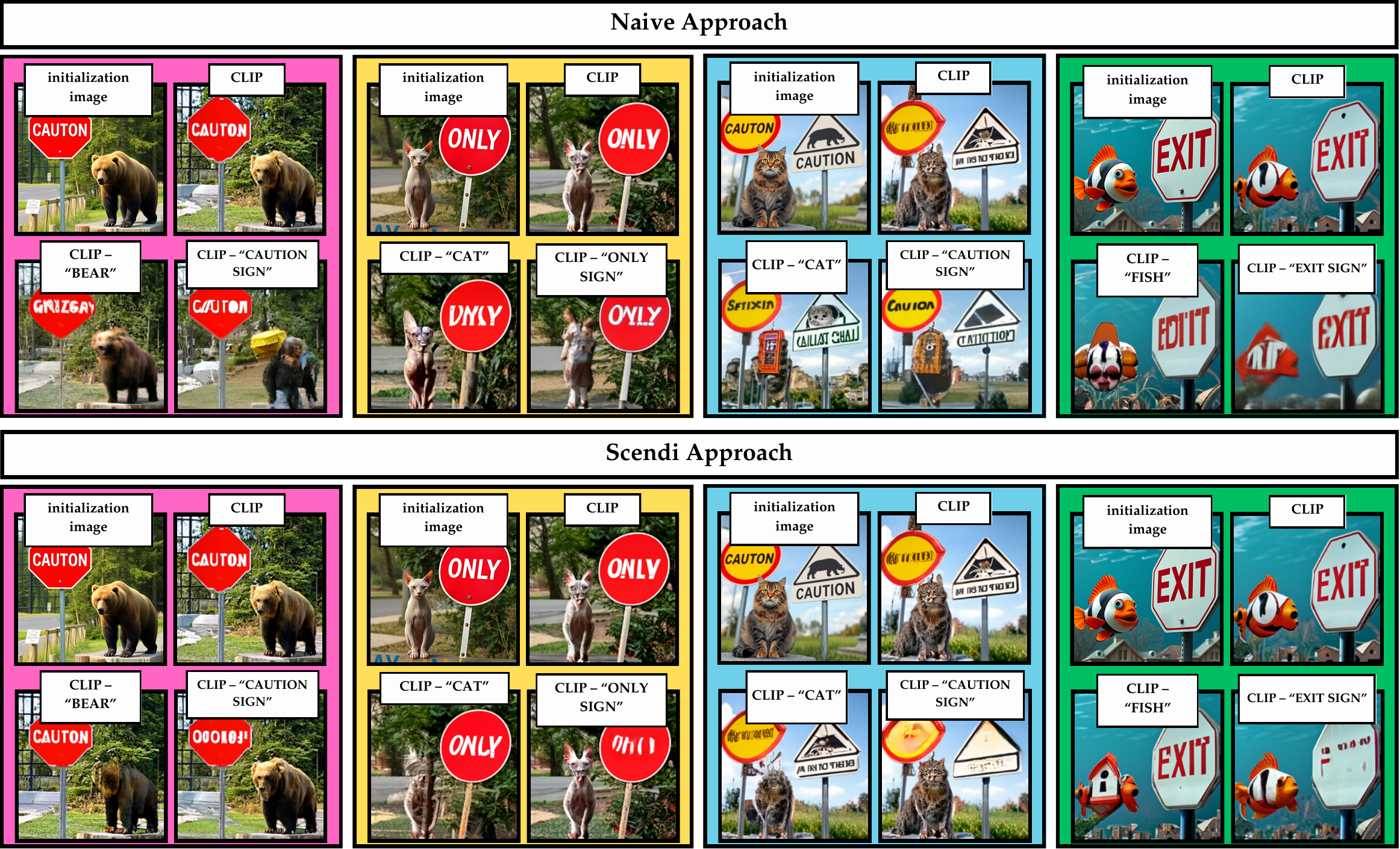}
    \caption{Comparison of samples generated by naive and SC-based decompositions of CLIP.}
    \label{fig:clip diffusion with naive}
\end{figure*}

Additionally, we performed CLIP-guided diffusion to visualize the contents of the corrected embeddings. The setup is illustrated in figure \ref{fig:diffusion clip naive diagram} and it is similar to the one described in the main text, except we do not use $\Gamma^*$ in the decomposition of CLIP. Figure \ref{fig:clip diffusion with naive} compares images generated using naive and SC-based decompositions. The naive method performs poorly, particularly when text overlays traffic signs, and often removes directions without preserving information about other underlying concepts in the images. In contrast, the SC-based decomposition preserves the structural and semantic information while successfully removing the undesired text directionality.

These results highlight the necessity of computing the correction matrix $\Gamma^*$ to effectively remove specific directions while preserving information about other concepts within the image embeddings.

\section{Additional Numerical Results}
\begin{figure}
    \centering
    \includegraphics[width=0.95\linewidth]{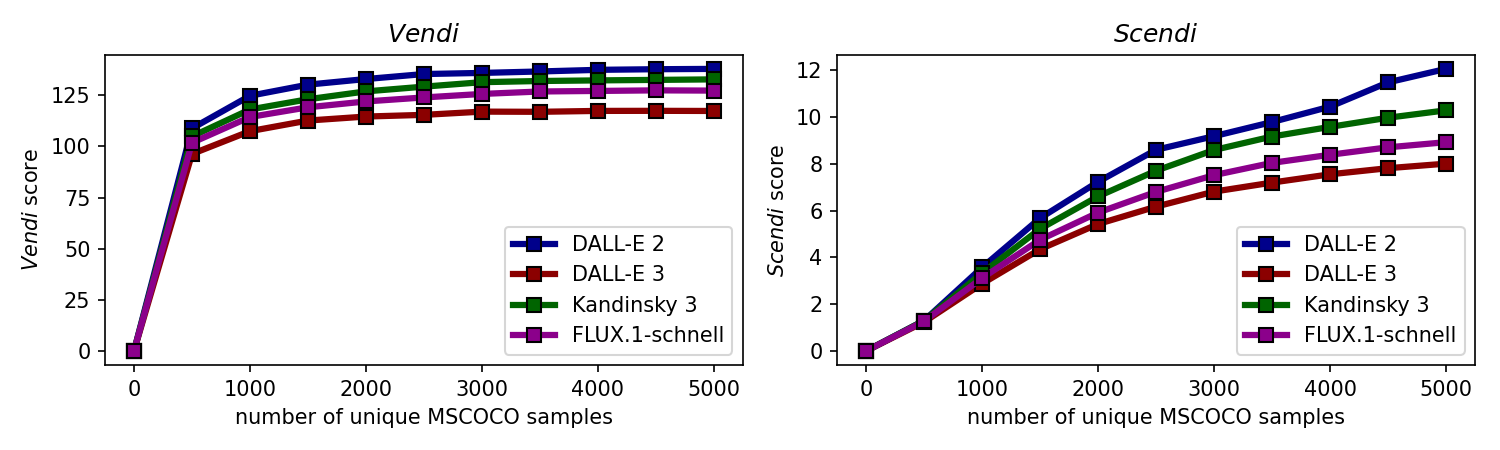}
    \caption{Comparison of different text-to-image models with Vendi-1.0 (generated image diversity) and Scendi (image generator diversity).}
    \label{fig:generative models comparison}
\end{figure}

We evaluated several text-to-image models using the Scendi metric to assess their performance. Figure \ref{fig:generative models comparison} summarizes our findings for DALL-E 2 \cite{ramesh2022hierarchicaltextconditionalimagegeneration}, DALL-E 3 \cite{dalle3}, Kandinsky 3 \cite{razzhigaev2023kandinsky}, and FLUX.1-schnell \cite{flux_2024}, tested on 5,000 MSCOCO \cite{lin2015microsoft} captions.

Our results demonstrate that the SC-Vendi metric correlates with the Vendi score, which measures the diversity of image generators. This suggests that when tested on the MSCOCO dataset, image diversity arises not only from the text prompts but also from the intrinsic properties of the generator itself.

\begin{figure}[h]
    \centering
    \includegraphics[width=\linewidth]{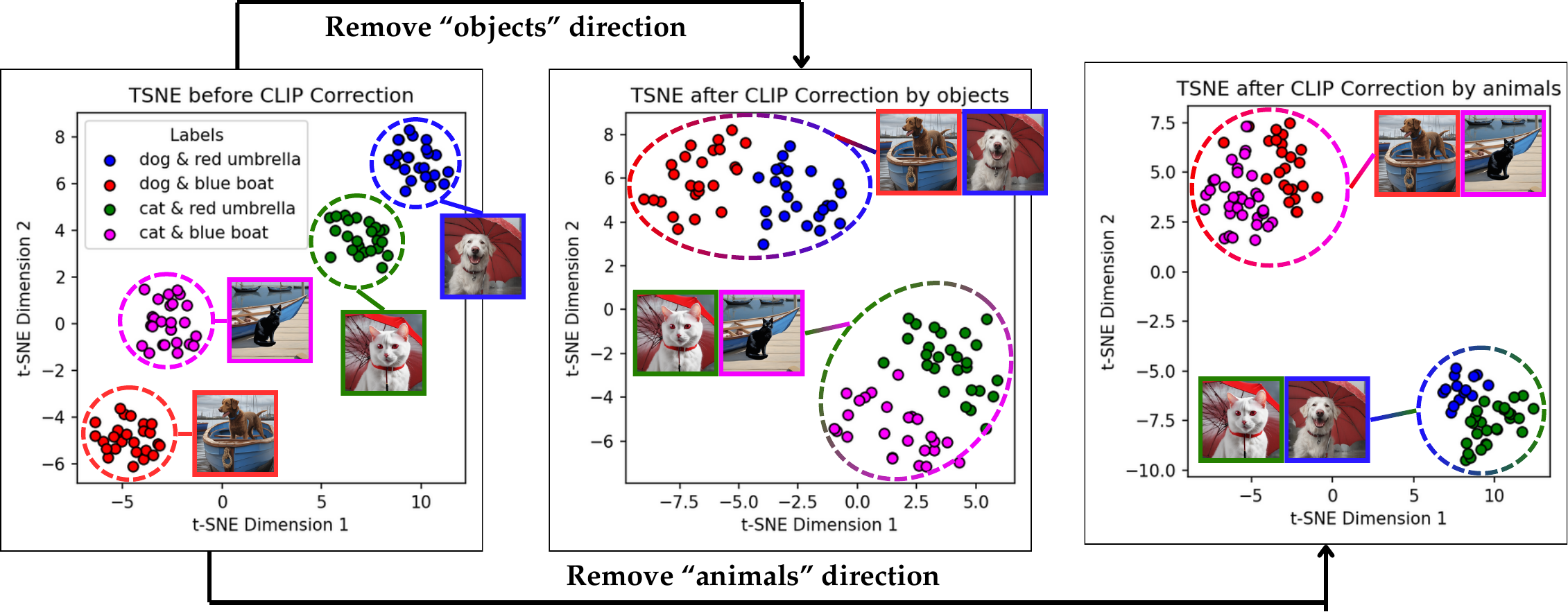}
    \caption{t-SNE plot of animals with objects dataset.}
    \label{fig:tsne animals objects}
    \vspace{-5mm}
\end{figure}

\end{document}